\documentclass{article}
\usepackage[T1]{fontenc}
\PassOptionsToPackage{dvipsnames}{xcolor}
\usepackage{iclr2027_conference,times}
\usepackage{amsmath,amssymb,amsfonts,amsthm,bm}
\usepackage{graphicx}
\usepackage[dvipsnames]{xcolor}
\usepackage{booktabs,multirow,threeparttable,makecell,colortbl}
\usepackage{algorithm,algpseudocode}
\usepackage{mathrsfs,pifont}
\usepackage{caption,subcaption}
\usepackage{enumitem,etoolbox,xspace}
\usepackage[most]{tcolorbox}
\usepackage{wrapfig,float,placeins,needspace}
\usepackage{url}
\usepackage{adjustbox}
\usepackage[breaklinks,hidelinks]{hyperref}

\makeatletter
\newcommand{\Rmnum}[1]{\expandafter\@slowromancap\romannumeral #1@}
\makeatother

\newcommand{\best}[1]{\cellcolor{black!25}#1}
\newcommand{\second}[1]{\cellcolor{black!10}#1}
\newcommand{\cmark}{$\checkmark$}
\newcommand{\xmark}{$\times$}

\newcommand{\blue}[1]{#1}

\newlength{\legacycolumnwidth}
\newcommand{\projecturl}{https://pilot-wam-2026.github.io/}

\makeatletter
\newcommand{\wrapfill}{%
  \par
  \ifnum\c@WF@wrappedlines>\@ne
    \null\vskip\c@WF@wrappedlines\baselineskip
    \global\c@WF@wrappedlines\z@
    \WF@finale
  \fi}
\newcommand{\FinishWrap}{%
  \par
  \ifnum\c@WF@wrappedlines>\z@
    \global\c@WF@wrappedlines\z@
    \WF@finale
  \fi}
\makeatother

\newsavebox{\TopPanelBox}
\newcount\TopPanelLines
\newdimen\TopPanelHeight
\newcommand{\TopWrap}[2]{%
  \setlength{\intextsep}{0pt}%
  \par
  \sbox{\TopPanelBox}{\begin{minipage}{#1}\captionsetup{type=figure}\input{#2}\end{minipage}}%
  \TopPanelHeight=\dimexpr\ht\TopPanelBox+\dp\TopPanelBox+8pt\relax
  \Needspace{\TopPanelHeight}%
  \TopPanelLines=\TopPanelHeight
  \divide\TopPanelLines by \baselineskip
  \advance\TopPanelLines by 1
  \begin{wrapfigure}[\the\TopPanelLines]{r}{#1}
    \usebox{\TopPanelBox}
  \end{wrapfigure}%
  \noindent\leavevmode\par\nobreak\kern-\baselineskip%
  \setlength{\intextsep}{12pt}%
}

\newcommand{\TableFont}{\scriptsize}

\usepackage{marvosym}
\definecolor{projectlinkblue}{rgb}{0.21,0.49,0.74}
\iclrfinalcopy
\title{Motion Chain of Thought: \\ 
Disentangling Motion Semantics from \\
Visual Appearance for Robotic Manipulation}
\author{%
Xiangkai Ma\textsuperscript{1}\quad
Xinyi Wang\textsuperscript{1}\quad
Junjie Wang\textsuperscript{2}\quad
Sheng Xu\textsuperscript{2}\quad
Mingyang Li\textsuperscript{3}\\
\bfseries
Yue Ma\textsuperscript{4}\quad
Han Zhang\textsuperscript{1}\quad
Yuzheng Zhuang\textsuperscript{5}\quad
Wenzhong Li\textsuperscript{1,\Letter}\quad
Zhihao Yuan\textsuperscript{5,}\thanks{\raggedright The corresponding authors are
Wenzhong Li (\mbox{\href{mailto:lwz@nju.edu.cn}{lwz@nju.edu.cn}}) and
Zhihao Yuan (\mbox{\href{mailto:zhihaoyuan@link.cuhk.edu.cn}{zhihaoyuan@link.cuhk.edu.cn}}).
Additional contact: Xiangkai Ma
(\mbox{\href{mailto:xiangkai.ma@smail.nju.edu.cn}{xiangkai.ma@smail.nju.edu.cn}}).}\\[3pt]
{\normalfont\small\textsuperscript{1}Nanjing University\quad
\textsuperscript{2}The Chinese University of Hong Kong, Shenzhen}\\
{\normalfont\small\textsuperscript{3}Tsinghua University\quad
\textsuperscript{4}Hong Kong University of Science and Technology}\\
{\normalfont\small\textsuperscript{5}Joy Future Academy, JD}%
}

\hypersetup{
  pdftitle={Motion Chain of Thought: Disentangling Motion Semantics from Visual Appearance for Robotic Manipulation},
  pdfauthor={Xiangkai Ma, Xinyi Wang, Junjie Wang, Sheng Xu, Mingyang Li, Yue Ma, Han Zhang, Yuzheng Zhuang, Wenzhong Li, Zhihao Yuan}
}

\makeatletter
\newcommand{\camerareadymaketitle}{%
  \begingroup
  \renewcommand{\@fnsymbol}[1]{%
    \ifcase##1\or\mbox{\Letter}\else\@ctrerr\fi}%
  \maketitle
  \endgroup
}
\makeatother

\begin{document}
\camerareadymaketitle
\begin{center}
\vspace{-30pt}
{\small\color{projectlinkblue}\url{https://pilot-wam-2026.github.io/}\par}
\vspace{5pt}
\includegraphics[width=\linewidth]{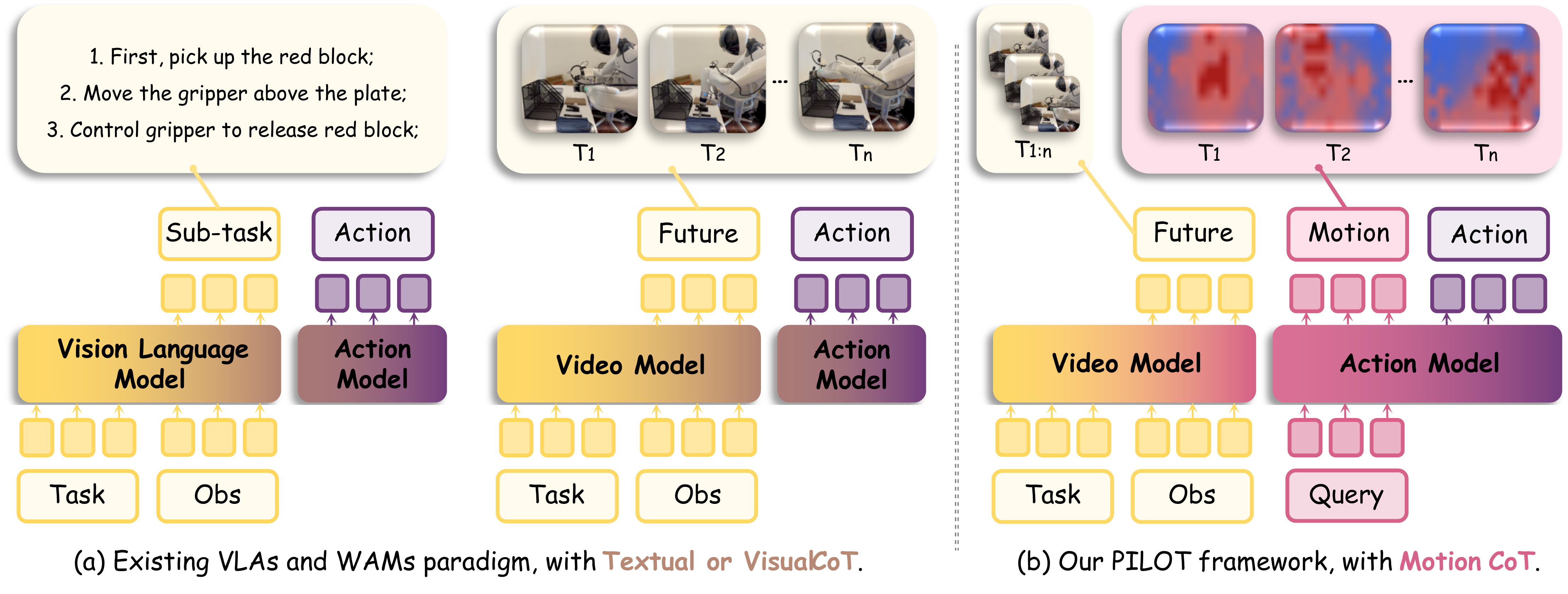}
\vspace{-10pt}
\captionsetup{type=figure}
\caption{
We propose \textbf{PILOT}, a framework that separates motion-semantic representation learning from trajectory refinement.
Representational Deduction (RD) supervises learnable Motion-CoT tokens derived from observation and instruction using future representations as training targets. Motion-CoT enables the action model to focus on fine-grained trajectory refinement while alleviating representational entanglement.%
}\label{fig:abstract}
\end{center}

\begin{abstract}
World Action Models (WAMs) aim to construct a unified architecture capable of understanding world state evolution and guiding action generation. 
However, existing visual branches focus on predicting static visual observation, rather than reflecting potential transition information that captures the evolution of world states under motion interactions. 
This leads to representational entanglement between high-level physical condition evolution and low-level action trajectory generation within the Action Model, creating a structural bottleneck while weakening the predictive capability of world evolution modeling for action generation. 
We propose PILOT (Physical Inference for Latent Optimized Trajectories), whose core Representational Deduction (RD) bridges this gap by integrating motion chain-of-thought (CoT) guidance as a native model capability. 
Specifically, RD aims to encourage the action branch to explicitly model potential state transition tokens, which are retained as CoT in the reasoning space to guide fine-grained motion trajectory. 
Experiments demonstrate that RD not only improves the success rate and generalization ability of WAMs in complex robotic manipulation tasks but also enhances the model's physical interpretability by decoupling high-level motion semantics from low-level trajectory details. 
Furthermore, the abundant state transition supervision signals introduced by RD effectively alleviate the sparse supervision in action generation, \blue{supporting data-efficient real-robot adaptation with reduced demonstration sets}.
\end{abstract}

\section{Introduction} \label{section:introduction}
World Models~\cite{hafner2023dreamerv3,bruce2024genie} and Embodied Intelligence~\cite{brohan2022rt1,brohan2023rt2,black2024pi0,intelligence2025pi05,kim2024openvla,xu2026reaction} are dedicated to ``understanding how the physical world evolves during human interaction'' and ``planning the action trajectory based on current observations,'' respectively. 
\TopWrap{\legacycolumnwidth}{figures/motiv}
The unification of these paradigms represents a significant step toward General Spatial Intelligence~\cite{yang2025thinking}. 
Recent World Action Models (WAMs)~\cite{bi2026motus,lyu2026lda,team2026motubrain,li2026wallwm,li2026lingbotva} approach this goal by leveraging world models to predict future states and introducing the Inverse Dynamics Module (IDM)~\cite{chen2025moto,liu2026lara,zhang2026defi} to convert these predictions into trajectory. 
However, future states are confined to the visual latent space constructed by frozen VAEs~\cite{pai2025mimic,cen2025rynnvla2,zhang2026dreamvla}. 
This structural bottleneck causes the joint optimization objective of WAMs to primarily target visual reconstruction over trajectory generation.
Guiding WAMs to focus on motion semantic representations to eliminate the visual bottleneck via an end-to-end framework remains a long-term goal.

Recent experiments~\cite{yuan2026fastwam,yang2026mantisvla,sun2026vlajepa} indicate that while explicit future prediction aids training, it provides limited foresight for trajectory generation during inference.
The effectiveness of WAMs stems from the future supervision to World Model, promoting the learning of physically meaningful visual priors. 
We attribute this to the fact that static visual representations at a single moment fail to reflect the evolution process of physical states under action interactions. 
Without explicit modeling of state transitions, directly generating actions through the diffusion process leads to representational entanglement, which is visualized in Fig.~\ref{fig:motiv}.
To address this, we propose that action model should simultaneously learn high-level motion intention structures and low-level manipulation details from the same inputs.

How to supervise the model to learn the motion semantic representations that contain state transition information? The existing WAM~\cite{ye2026dreamzero,gao2026dreamdojo,cen2025worldvla,feng2025vidar} can precisely predict what will happen in the future, but still doesn't know how to do it. 
Most relevant works are Latent Action Models (LAMs), which typically capture pixel-level differences and reconstruct future frames from the current frame via IDM~\cite{ye2025lapa,bu2025univla,routray2025vipra}. 
However, this kind of visual-aware-only supervision, fails to prompt the model to understand what action conditions led to the variation in the visual state. 
Instead of deriving latent motions directly from consecutive frames, we employ them exclusively as supervision, guiding the model to summarize physical state transitions and motion semantics within learnable tokens.
We propose not to pursue to capture motion-related state transition semantics from pixel variations between two frames, but rather to use variations in physical spatial states as supervisions.

In this paper, we propose Physical Inference for Latent Optimized Trajectories \textbf{(PILOT)}, an end-to-end framework grounded in the Representational Deduction \textbf{(RD)} mechanism that bridges the gap between static visual prediction and dynamic action manipulation by establishing motion semantic CoT reasoning as an innate capability of WAMs. 
Architecturally, PILOT decouples the World Model into understanding and generation pipelines to substitute explicit future image prediction with the Motion-CoT during inference, thereby accelerating deployment speed while improving both simulation and real-world.
Specifically, the Action Model derives high-level Motion-CoT tokens from the current observation, instruction, and proprioceptive state; future observations are used only as training supervision for learning predictive state-transition representations.
Subsequently, we freeze the pre-trained VJEPA-as-Encoder~\cite{assran2025vjepa2,mur2026vjepa21,maes2026leworldmodel} to capture deterministic features tightly coupled with action conditions, and train a Causal Dynamics Engine \textbf{(CDE)} to predict future spatial positions, trajectories, shapes, and interaction relationships conditioned on motion semantic CoT, rather than probabilistic underlying visual rendering details. 
Meanwhile, these motion semantics are retained as context tokens in the reasoning space, allowing the subsequent diffusion process to concentrate on refining fine-grained end-effector trajectories and achieve representational decoupling. 
This brings an additional benefit: Motion-CoT serves as an extra supervision signal, alleviating the sparse supervision problem faced by the action model due to the low information density of action signals.
Our contributions can be summarized as follows. 
\begin{itemize}
    \item We propose PILOT, a novel WAM framework that decouples motion intention from trajectory generation via a Representational Deduction mechanism, and internalizes latent state transition information as Motion CoT to alleviate representational entanglement.
    \item We introduce a Causal Dynamics Engine for future-representation supervision and asymmetric attention in Action-Perceiver that prevents noised actions from influencing motion-semantic queries, enabling action inference without future pixel reconstruction.
    \item PILOT achieves 97.9\%, 80.3\%, and 59.75\% success on LIBERO, LIBERO-Plus, and RoboCasa-GR1, respectively, and 82.5\% average success on the eight standard real-world tasks. A separate pipeline-timing comparison reports 91.6\% lower latency for action-only inference than predict-then-act inference.
\end{itemize}

\section{Related Work}\label{section:relatedwork}
\paragraph{World Models for Robotic Manipulations}
Recently, WAMs~\cite{cen2025worldvla,li2026wallwm,li2026lingbotva,cen2025rynnvla2,zhang2026dreamvla,zhang2026imagewam} have emerged as a promising paradigm that leverages world model to predict future, thereby guiding motion manipulation. 
Existing video-model-based WAMs~\cite{kim2026cosmospolicy} generally follow two paradigms. 
The serial-paradigm first predicts future observation and then reconstructs actions via an inverse dynamic~\cite{pai2025mimic,chen2025moto,zhou2024robodreamer}, while the parallel-paradigm jointly models the visual-action space through shared attention mechanism (MoT)~\cite{bi2026motus,lyu2026lda,yuan2026fastwam} or joint denoising~\cite{ye2026dreamzero,gao2026dreamdojo}. 
Unlike generative pixel reconstruction strategies, joint embedding predictive frameworks (JEPA)~\cite{maes2026leworldmodel,assran2025vjepa2,mur2026vjepa21} focus on predicting dynamic features such as object spatial positions and trajectories. 
Some studies~\cite{jha2026reconstruction,su2026wog} incorporated JEPA-as-encoder to obtain abundant representations. 
Most relevant to our work is VLA-JEPA~\cite{sun2026vlajepa}, which introduces JEPA to extract latent states, addressing the challenges of pixel reconstruction bias and future information leakage.
\blue{Related JEPA-based WAMs provide predictive guidance through visual-transition or stage prediction~\cite{lin2026jepawam1,liu2026jepawam4}, visual-instruction embeddings~\cite{liu2026jepawam3}, or future-scene representations jointly predicted with trajectories~\cite{wang2026jepawam2}.}

\blue{However, purely visual supervision fails to provide state transition information. 
This inherent discrepancy causes the deviation of co-training from the motion semantics, resulting in fragile transferability and inefficient fine-tuning; 
conversely, completely removing visual supervision and relying solely on sparse action supervision leads to representation collapse of video model. 
To address this, we propose Representational Deduction, which bridges the aforementioned gap by incorporating additional motion semantics and physical state transition information.}

\paragraph{Latent Action Learning for Robotics}
LAMs~\cite{routray2025vipra,liu2026lara,bruce2024genie} aim to pre-train backbones on large-scale human videos lacking action labels, thereby establishing prior knowledge of latent motion. 
Existing approaches~\cite{ye2025lapa,bu2025univla,chen2025moto,zhang2026defi,yang2026como} typically align vision and manipulation representations using Q-former~\cite{li2023blip2} and unified codebook~\cite{zhang2026clap,ma2026unifying,chen2025moto,chen2026lawam} to integrate robotic manipulation into the Next-Token-Prediction framework.
Despite the promise of LAMs, their performance is hindered by noise from camera-induced background variations and future frame leakage, which encourages trivial encoding of future frames rather than learning state transitions~\cite{sun2026vlajepa}. 
These issues cause structural bottlenecks in learning control-oriented latent actions. Recent works mitigate this via latent constraints~\cite{luo2026beingh07} or multi-encoder-query~\cite{su2026wog}. 
However, the LAMs inherently aligns latent actions with visual space rather than actionable control signals. 
PILOT fundamentally avoids these defects by ensuring that future visual information is used exclusively as a supervision target, not as a source for capturing latent motion representations, thereby guaranteeing that model focuses on dynamic representations independent of camera motion and background changes.

\newcommand{\modelname}{PILOT}
\newcommand{\modelfull}{Physical Inference for Latent Optimized Trajectories}
\newcommand{\RDname}{Representational Deduction}
\newcommand{\rdname}{representational deduction}
\newcommand{\RDabbr}{RD}
\newcommand{\rdabbr}{RD}
\newcommand{\CDEname}{Causal Dynamics Engine}
\newcommand{\CDEabbr}{CDE}

\section{Method}
\label{sec:method}
\begin{figure}[!t]
\input{figures/method}
\end{figure}

As illustrated in Fig.~\ref{fig:method1}, \modelname{} combines a Cosmos-Predict2.5 world model~\citep{nvidia2025cosmospredict25} for vision--language context and auxiliary future prediction, a Perceiver-style Action Model~\citep{jaegle2021perceiver} for motion-semantic queries and flow-matching action decoding, and an \RDabbr{} branch for VJEPA2-AC future-representation supervision~\citep{assran2025vjepa2}.

\subsection{Problem formulation and notation.}
At time $t$, the policy receives an RGB observation $o_t$, instruction $\ell$, and proprioceptive state $s_t\in\mathbb{R}^{d_s}$, and predicts an $H$-step action chunk $\bm{a}_t=(a_t,\ldots,a_{t+H-1})\in\mathbb{R}^{H\times d_a}$. Training additionally uses a future observation $o_{t+\Delta}$ from the same trajectory. With context $\bm{c}=\mathrm{WM}_{\theta}(o_t,\ell)$, policy is
\begin{equation}
  p_{\Theta}(\bm{a}_t\mid o_t,\ell,s_t)
  = \int p_{\Theta}(\bm{a}_t\mid \bm{m},\bm{c},s_t)\, p_{\Theta}(\bm{m}\mid \bm{c},s_t)\,\mathrm{d}\bm{m},
  \label{eq:factor}
\end{equation}
where $\bm{m}=f_{\Theta}(\bm{c},s_t)\in\mathbb{R}^{K\times d}$ is deterministic, so its conditional distribution is a point mass. The action decoder uses motion semantics, visual context, and proprioceptive state together.

\subsection{Motion-CoT from Learnable Queries}\label{sec:method:motion}
\paragraph{Vision-language context.}
In the comprehension pipeline, we repurpose Cosmos-Predict2.5 as a vision-language encoder. The video tokenizer encodes observation $o_t$ into clean latents and the text encoder embeds instruction $\ell$; a single conditional forward pass yields the context sequence
\begin{equation}
  \bm{c} \;=\; \mathrm{WM}_{\theta}(o_t,\ell)\;\in\;\mathbb{R}^{N_c\times d_c},
  \label{eq:context}
\end{equation}
where $\mathrm{WM}_{\theta}$ denotes the trainable world-model parameters, and $N_c$ and $d_c$ are the number and dimension of context tokens. The pre-trained video prior in $\mathrm{WM}_{\theta}$ provides visual and physical cues for downstream action generation.

\paragraph{Learnable motion-semantic queries.}
We introduce $K$ learnable queries $\bm{Q}\in\mathbb{R}^{K\times d_h}$ in a Perceiver-style flow-matching head, with initialization and internal width $d_h=768$. The final slot projection produces Motion-CoT outputs of width $d=1024$. Motion-CoT denotes these learned semantic slots: their indices have no prescribed temporal ordering and are distinct from the $H$ action steps.
At a sampled flow time $\tau$, the noised action chunk $\bm{a}^{\tau}_t=(1-\tau)\,\bm{\epsilon}+\tau\,\bm{a}_t$ is embedded into action tokens $\bm{A}^{\tau}\in\mathbb{R}^{H\times d_h}$, where $\bm{\epsilon}\!\sim\!\mathcal{N}(\bm{0},\bm{I})$.
The state $s_t$ is mapped into a state token $\bm{S}\in\mathbb{R}^{1\times d_h}$, and the latent sequence $[\bm{S};\bm{Q};\bm{A}^{\tau}]$ attends to $\bm{c}$ through a stack of $L$ blocks.

\vspace{-5pt}
\paragraph{Causally-decoupled attention mechanism.}
To prevent motion semantics from absorbing diffusion noise, we decouple the attention inside each Perceiver block. Denoting the non-action latents as $\bm{Z}=[\bm{S};\bm{Q}]$, each block applies
\begin{align}
  \bm{Z}    &\leftarrow \bm{Z} + \mathrm{Attn}\!\big(\bm{Z},\,[\bm{c};\bm{Z}]\big),
  \label{eq:attn-query}\\
  \bm{A}^{\tau} &\leftarrow \bm{A}^{\tau} + \mathrm{Attn}\!\big(\bm{A}^{\tau},\,
                 [\bm{c};\bm{Z};\bm{A}^{\tau}]\big),
  \label{eq:attn-action-1}
\end{align}
where $\bm{Z}$ attends to itself and visual context, while action tokens also attend to $\bm{Z}$. Here, ``causally-decoupled'' describes the directed information flow: queries condition actions, while reverse attention from noised actions to queries is blocked.
Time-conditioned AdaLN is applied only to action tokens, and $\bm{Z}$ uses plain layer normalization to remain invariant to the flow time $\tau$.
After the final block and the $d_h\to d$ output projection, we read out the query slots as the motion-semantic context:
\begin{align}
  \bm{O}=\mathrm{Act.Per.}(\bm{Z},\bm{A}^{\tau},\bm{c};\tau)\;\in\;\mathbb{R}^{(1+K+H)\times d},
  \qquad
  \bm{m}=\bm{O}[:,\,1\!:\!K\!+\!1,\, :]\;\in\;\mathbb{R}^{K\times d}.
  \label{eq:attn-action-2}
\end{align}

\paragraph{Flow-matching action decoding.}
Action tokens after the last block are projected back to the action space and supervised with the conditional flow-matching velocity objective:
\begin{equation}
  \mathcal{L}_{\mathrm{act}}
   \;=\;
   \mathbb{E}_{\tau,\bm{\epsilon}}
   \big\|\, \mathrm{Proj}_{\mathrm{act}}\!\big(\bm{O}[:,\,K\!+\!1:,\, :]\big)
            -(\bm{a}_t-\bm{\epsilon}) \,\big\|^2,
  \label{eq:loss-action}
\end{equation}
where $\mathrm{Proj}_{\mathrm{act}}$ is an MLP mapping action tokens to the action dimension $d_a$. 
At inference, Euler integration starts from Gaussian noise, conditioned on $\bm{m}$, $\bm{c}$, and $s_t$.

\subsection{\RDname{} with a \CDEabbr{}}\label{sec:method:rs}
RD trains $\bm{m}$ to support prediction of a future representation.
The \CDEabbr{} models a directed current-to-future mapping conditioned on motion semantics and robot state.

\paragraph{Frozen state representations.}
Given the current and future observations $(o_t,o_{t+\Delta})$, frozen VJEPA2-AC encoder $E_{\psi}$ produces per-frame patch representations $\bm{r}_t\textit{=}E_{\psi}(o_t),\bm{r}_{t+\Delta}\textit{=}E_{\psi}(o_{t+\Delta})$, 
where each representation lies in $\mathbb{R}^{P\times d_r}$, with $P$ patch tokens and token dimension $d_r$. 
The image encoder provides the representation targets; action conditioning belongs to the pretrained predictor. RD uses the encoded future observation only as a training target, while queries are computed from current context and state.

\paragraph{Causal dynamics engine.}
The \CDEabbr{} $G_{\xi}$ is initialized from the pre-trained VJEPA2-AC predictor and remains trainable. 
Instead of using the original action and state inputs of the predictor, we condition the \CDEabbr{} on learned motion semantics and robotic state. 
Two lightweight linear projectors first map the motion semantics and proprioceptive state to encoder feature dimensionality:
\begin{equation}
  \tilde{\bm{m}}=\bm{W}_m\,\bm{m} \in \mathbb{R}^{K\times d_r},
  \qquad
  \tilde{\bm{s}}=\bm{W}_s\,s_t \in \mathbb{R}^{1\times d_r}.
  \label{eq:project}
\end{equation}
Inside the CDE, these conditioning tokens are further mapped from $d_r\textit{=}1408$ to the predictor's internal width, $d_p\textit{=}1024$, and concatenated with the projected current-frame patch features. The predictor outputs are then mapped back to $d_r\textit{=}1408$ to obtain the future patch-feature grid:
\begin{equation}
  \hat{\bm{r}}_{t+\Delta}
   \;=\; G_{\xi}\big(\bm{r}_t \,;\, \tilde{\bm{m}},\, \tilde{\bm{s}}\big)
   \;\in\;\mathbb{R}^{P\times d_r}.
  \label{eq:evolve}
\end{equation}
Conditioning tokens are prepended to current-frame patches, and one $P\times d_r$ future representation is returned. The $K$ query slots jointly condition this prediction rather than represent successive future states.

\paragraph{\rdname{} loss.}
We supervise the \CDEabbr{} by matching its prediction to the frozen encoder representation of the ground-truth future frame:
\begin{equation}
  \mathcal{L}_{\mathrm{\rdabbr{}}}
   \;=\;
   \mathrm{SmoothL1}\big(\hat{\bm{r}}_{t+\Delta},\, \bm{r}_{t+\Delta}\big).
  \label{eq:loss-rs}
\end{equation}
Gradients from $\mathcal{L}_{\mathrm{\rdabbr{}}}$ propagate through the \CDEabbr{} and the projector $\bm{W}_m$ to the learnable queries, encouraging $\bm{m}$ to encode transition information that is predictive of future VJEPA2-AC representations. 
This objective supplies transition-oriented representation supervision alongside the action loss and the auxiliary future-frame objective.

\begin{figure}[!t]
\input{figures/expr5}
\end{figure}

\subsection{World-Model Branch and Training Objective}\label{sec:method:wm}

\paragraph{Future-frame latent prediction.}
In the generation pipeline, the world model retains its generative capability through an auxiliary future-frame objective. 
We encode the short clip $[\,o_t,\,o_{t+\Delta}\,]$ into clean video latents, using the current-frame latents as conditions and the future-frame latents as prediction targets. 
Flow-matching noise is applied only to the target latents, and the world model is trained to predict the corresponding target velocity:
\begin{equation}
  \begin{aligned}
  \mathcal{L}_{\mathrm{wm}}
   &= \mathbb{E}_{\tau,\bm{\epsilon}}
   \big\|\, \mathrm{WM}_{\theta}(\bm{z}^{\tau};\,o_t,\ell,\tau) -(\bm{\epsilon}-\bm{z}^{\ast}) \,\big\|^2_{\odot\,\bm{M}_{\bm{z}^{\tau}}},
  \end{aligned}
  \label{eq:loss-wm}
\end{equation}
where $\mathrm{WM}_{\theta}$ denotes the trainable world-model parameters, $\bm{z}^{\ast}$ are the clean target latents, $\bm{z}^{\tau}$ is their noised version, and $\bm{M}_{\bm{z}^{\tau}}$ restricts the loss to future-frame positions. 
This objective preserves the instruction-grounded future prior of the pre-trained generator, while the generation pipeline is disabled during inference to avoid the visual bottleneck.

\paragraph{Overall objective.}
The three branches are trained jointly with a single weighted objective:
\begin{equation}
  \mathcal{L}
   \;=\;
   \mathcal{L}_{\mathrm{act}}
   \;+\; \lambda_{\mathrm{wm}}\,\mathcal{L}_{\mathrm{wm}}
   \;+\; \lambda_{\mathrm{\rdabbr{}}}\,\mathcal{L}_{\mathrm{\rdabbr{}}},
  \label{eq:loss-total}
\end{equation}
where $\lambda_{\mathrm{wm}}$ and $\lambda_{\mathrm{\rdabbr{}}}$ balance the auxiliary future-frame and \RDname{} losses against the action objective. 
In practice, the VJEPA2-AC encoder is frozen; the world model and \CDEabbr{} are fine-tuned from their pre-trained weights; and the Action Model together with the two projectors is trained from scratch.

\paragraph{Inference.}
At test time, the world model encodes $(o_t,\ell)$ into $\bm{c}$, the state-conditioned queries produce $\bm{m}$, and the flow-matching head generates the action chunk from $(\bm{m},\bm{c},s_t)$.
The \RDabbr{} branch and future-frame prediction serve only as training-time supervision and incur no additional inference cost; the generative decoder can be optionally invoked for future-frame visualization.
\FinishWrap

\section{Experiments} \label{section:experiments}

\begin{samepage}
\begin{wrapfigure}{r}{0.5\textwidth}
  \input{figures/expr4}
\end{wrapfigure}
\paragraph{Implementation Details}
\label{sec:exp:impl}
Our codebase is built upon the StarVLA project~\citep{ye2026starvla}.
\modelname{} is implemented with three cooperative branches.
The \textbf{World-Model branch} builds on a pre-trained Cosmos-Predict2.5 video diffusion transformer, which simultaneously serves as a visual-language encoder (providing a $196\!\times\!2048$ context) and a trainable latent future-frame generator.
The \textbf{Action-Model branch} uses $K=64$ learnable query tokens with initialization and internal width $d_h=768$, projected to Motion-CoT outputs of width $d=1024$, and a DiT-based flow-matching action decoder with 20 denoising steps.
The \textbf{\RDname{} branch} employs a frozen VJEPA2-AC encoder that produces current and future patch-level representations ($256\!\times\!1408$), and a lightweight \CDEname{} that predicts future representations conditioned on motion semantics and proprioceptive state.

\end{samepage}

\subsection{Simulation Benchmarks}
\label{sec:exp:sim}

\paragraph{LIBERO.}
LIBERO~\cite{liu2023libero} is a tabletop manipulation benchmark with a Franka Panda robot, featuring four task suites of increasing complexity: Spatial, Object, Goal, and Long-horizon.

\begin{wraptable}{r}{0.5\textwidth}
  \centering
\caption{
\textbf{Evaluation on LIBERO~\citep{liu2023libero}.}
Success rates (\%) on Spatial, Object, Goal, and Long-horizon suites. Size denotes parameter count (B: billions); $^\dagger$ denotes trainable parameters for PILOT, with its parameter breakdown in Appendix~\ref{sec:appendix:parameters}. Avg.\ averages the four suite scores. Baselines are published results. VLA/LAM/WAM group vision-language-, latent-, and world-action models.
}
\label{tab:libero}
\TableFont
\begingroup
\begingroup\TableFont
\tabcolsep=1.8pt
\renewcommand\arraystretch{1.0}
\noindent\begin{tabular}{c l|c|cccc|c}
\toprule
Group & Method & Size & Spatial & Object & Goal & Long & \textbf{Avg.} \\
\midrule
\multirow{4}{*}{VLA} & \blue{CoT-VLA} & \blue{7B} & \blue{87.5} & \blue{91.6} & \blue{87.6} & \blue{69.0} & \blue{83.9} \\
 & $\pi_{0.5}$ & 3.5B & \best{98.8} & 98.2 & \second{98.0} & 92.4 & 96.9 \\
 & GR00T-N1.6 & 3.3B & 97.5 & 98.5 & 97.5 & 94.4 & 97.0 \\
 & OpenVLA-OFT & 7B & 97.6 & 98.4 & 97.9 & 94.5 & 97.1 \\
\midrule
\multirow{4}{*}{LAM} & LAPA$^*$ & 7B & 73.8 & 74.6 & 58.8 & 55.4 & 65.7 \\
 & UniVLA & 7B & 96.5 & 96.8 & 95.6 & 92.0 & 95.2 \\
 & Mantis & 5.8B & \best{98.8} & 99.2 & 94.4 & 94.2 & 96.7 \\
 & VLA-JEPA & 3B & 96.2 & 99.6 & 97.2 & 95.8 & 97.2 \\
\midrule
\multirow{4}{*}{WAM} & F1 & 4B & \second{98.2} & 97.8 & 95.4 & 91.3 & 95.7 \\
 & Fast-WAM & 6B & \second{98.2} & \best{100.0} & 97.0 & 95.2 & 97.6 \\
 & Motus & 8B & 96.8 & \second{99.8} & 96.6 & \best{97.6} & \second{97.7} \\
 & \textbf{\modelname{} (Ours)} & \textbf{2.5B}$^\dagger$ & 97.1 & 99.2 & \best{98.1} & \second{97.2} & \best{97.9} \\
\bottomrule
\end{tabular}\par
\endgroup
\endgroup

\end{wraptable}

As shown in Table~\ref{tab:libero}, \modelname{} achieves \textbf{97.9\%} average success rate with approximately 2.5B trainable parameters. This success rate surpasses all baselines, including Motus (97.7\%) and $\pi_{0.5}$ (96.9\%).
Notably, \modelname{} excels on the challenging Goal (98.1\%) and Long-horizon (97.2\%) suites, where understanding the \emph{intent} of manipulation is critical for long-horizon planning.
This validates our central thesis: decoupling motion semantics from trajectory generation via \RDname{} effectively alleviates representational entanglement.

\paragraph{LIBERO-Plus.}
As reported in Table~\ref{tab:libero_plus}, \modelname{} achieves \textbf{80.3\%} condition-count-weighted success rate, exceeding PokeVLA (79.3\%) by 1.0 percentage point in the displayed comparison. 
The improvement is particularly pronounced on Robot (70.0\% vs.\ 46.1\%) and Layout (82.0\% vs.\ 77.2\%), suggesting that the motion-semantic tokens learned by \RDname{} encode \emph{transition dynamics} rather than superficial appearance, making the policy inherently more robust to structural perturbations.

\begin{table}[H]
\centering
\caption{
\textbf{Evaluation on LIBERO-Plus~\citep{fei2025libero}.}
Success rates (\%) under Camera, Robot, Language, Light, Background, Noise, and Layout perturbations. Policies trained on standard LIBERO are evaluated by zero-shot transfer, testing robustness beyond the training distribution. VLA/LAM/WAM denote vision-language-, latent-, and world-action models; Size is parameter count, with $^\dagger$ denoting trainable parameters for PILOT (Appendix~\ref{sec:appendix:parameters}), and Avg.\ the condition-count-weighted mean recomputed from the displayed category scores. The appendix specifies the aggregation weights.
}
\label{tab:libero_plus}
\TableFont
\begingroup
\begingroup\TableFont
\tabcolsep=2pt
\renewcommand\arraystretch{1}
\noindent\begin{tabular*}{\linewidth}{@{\extracolsep{\fill}}c l|c|ccccccc|c@{}}
\toprule
Group & Method & Size
& Cam. & Rob. & Lang. & Light & Back. & Noise & Layout & \textbf{Avg.} \\
\midrule
\multirow{5}{*}{VLA} & \blue{ECoT} & \blue{7B} & \blue{0.3} & \blue{26.8} & \blue{40.2} & \blue{42.6} & \blue{16.4} & \blue{10.2} & \blue{36.9} & \blue{24.3} \\
& Spatial Forcing & 7B & 20.1 & 13.4 & 40.9 & 29.1 & 33.4 & 25.7 & 39.3 & 28.5 \\
 & $\pi_{0}$-FAST & 3.3B & 65.1 & 21.6 & 61.0 & 73.2 & 73.2 & 74.4 & 68.8 & 61.6 \\
 & OpenVLA-OFT & 7B & 56.4 & 31.9 & 79.5 & 88.7 & \best{93.3} & 75.8 & 74.2 & 69.6 \\
 & PokeVLA & 1.22B & \best{84.7} & 46.1 & \best{84.8} & \best{94.6} & 82.6 & \best{89.8} & \second{77.2} & \second{79.3} \\
\midrule
LAM & UniVLA & 7B & 1.8 & \second{46.2} & 69.6 & 69.0 & 81.0 & 21.2 & 31.9 & 42.9 \\
\midrule
\multirow{3}{*}{WAM} & WorldVLA & 7B & 0.1 & 27.9 & 41.6 & 43.7 & 17.1 & 10.9 & 38.0 & 25.0 \\
 & Fast-WAM & 6B & 16.4 & 44.5 & 68.9 & 78.2 & 53.7 & 37.7 & 60.7 & 50.0 \\
 & \textbf{\modelname{} (Ours)} & \textbf{2.5B}$^\dagger$ & \second{76.0} & \best{70.0} & \second{84.0} & \second{92.0} & \second{85.0} & \second{78.0} & \best{82.0} & \best{80.3} \\
\bottomrule
\end{tabular*}\par
\endgroup
\endgroup
\end{table}

\begin{table}[H]
\centering
\caption{\textbf{RoboCasa-GR1: Category success rates (\%).}
Comparison of \blue{13} methods across 24 tasks; parentheses give task counts. PILOT uses the released step-340000 checkpoint, with 50 rollouts/task and 717/1200 successes. Category scores average their constituent tasks; overall scores weight all 24 tasks equally. PnP denotes pick-and-place; other groups denote novel source locations. \colorbox{black!25}{Best} and \colorbox{black!10}{second-best} values are shaded per row; ties share a shade.
D4D: DiT4DiT; UWM: UWM-XL; QG: QwenGR00T; VP: VP-VLA; Phys.: PhysBrain-4B; Lang.: LangForce.
\blue{TwinB.: TwinBrainVLA (Qwen3-VL); StarVLA: StarVLA-alpha Generalist.}
Full PILOT and nine-baseline task results are in Appendix~\ref{sec:appendix:robocasa_results}.}
\label{tab:robocasa}
\TableFont
\renewcommand\arraystretch{1}
\begingroup\tabcolsep=2pt
\noindent\begin{tabular*}{\linewidth}{@{\extracolsep{\fill}}l|cccccccccccc|c@{}}
\toprule
\textbf{Category}
& \makecell{GR00T\\N1.5}
& \makecell{GR00T\\N1.6}
& D4D
& UWM
& QG
& LDA
& VP
& Phys.
& Lang.
& \blue{TwinB.}
& \blue{StarVLA}
& \blue{ABot-M0}
& \TableFont\textbf{\modelname{}} \\
\midrule
PnP Close (6)
& 45.3 & 24.2 & 50.3 & 30.7 & 50.3 & 56.3 & 54.3 & 52.7 & 55.7
& \second{\blue{56.67}} & \blue{54.00} & \best{\blue{61.67}} & 47.67 \\

Cuttingboard (5)
& 46.4 & 56.9 & 57.6 & 17.4 & 52.8 & \second{64.2} & 60.8 & 52.0 & 53.2
& \blue{59.20} & \blue{57.20} & \blue{61.60} & \best{70.00} \\

Placemat (4)
& 45.5 & 51.9 & 39.0 & 10.0 & 38.0 & 47.8 & \second{54.5} & 48.0 & 48.0
& \blue{46.50} & \best{\blue{59.00}} & \blue{51.00} & \second{54.50} \\

Tray (5)
& 48.8 & 55.1 & 46.4 & 17.2 & 39.2 & 53.4 & 46.0 & 46.0 & 47.2
& \blue{49.60} & \blue{53.80} & \second{\blue{56.80}} & \best{73.20} \\

Plate (4)
& 56.5 & 57.6 & 60.0 & 16.3 & 58.5 & 53.0 & 53.5 & \second{62.5} & 58.5
& \blue{60.00} & \best{\blue{65.00}} & \blue{58.00} & 53.50 \\
\midrule
\textbf{Average}
& 48.2 & 47.6 & 50.8 & 19.3 & 47.8 & 55.4 & 53.8 & 52.0 & 52.6
& \blue{54.58} & \blue{57.29} & \second{\blue{58.25}} & \best{59.75} \\
\bottomrule
\end{tabular*}\par\endgroup
\end{table}

\begin{wrapfigure}{r}{0.5\textwidth}
  \input{figures/expr3}
\end{wrapfigure}
\paragraph{RoboCasa-GR1.}
As shown in Table~\ref{tab:robocasa}, the released step-340000 \modelname{} checkpoint achieves \textbf{59.75\%} success (717/1200 episodes) over 24 tasks with 50 rollouts each, exceeding the strongest displayed baseline, ABot-M0 (58.25\%), by 1.50 percentage points.
Dark gray marks its best overall score and leading Cuttingboard (70.00\%) and Tray (73.20\%) results; light gray marks its second-best Placemat score (54.50\%), tied with VP-VLA.
The largest category gain is on Tray (73.20\% vs.\ 56.80\% for ABot-M0), an improvement of 16.40 points. Within Tray, success reaches 84\% for Pot, 82\% for Cardboardbox, and 80\% for Plate.
PnP Close and Plate average 47.67\% and 53.50\%, respectively. Cuttingboard-to-Pan and Cuttingboard-to-Pot reach 90\% and 92\%, while Placemat-to-Basket reaches 84\%.
Against the nine baselines with complete task-level results in Appendix~\ref{sec:appendix:robocasa_results}, \modelname{} ranks first on 11 tasks and second on one. The overall score weights each of the 24 tasks equally. Checkpoint-specific evaluation settings and public logs are documented in Appendix~\ref{sec:appendix:exp_details}.

\subsection{Real-World Evaluation}
\label{sec:exp:real}

\paragraph{Standard Tasks.}
Each of the eight standard desk tasks is evaluated over 50 rollouts.
Table~\ref{tab:realworld} compares \modelname{} with $\pi_0$, GR00T-N1.7, and $\pi_{0.5}$.
\modelname{} leads on six tasks: Pen, Eraser, Charger, Pencil Case, Stapler, and Trash.
Gains over $\pi_{0.5}$ reach 32 points on Eraser, 18 on Stapler and Trash, 16 on Pencil Case, and 12 on Charger.
Compared with $\pi_0$, \modelname{} improves all eight standard tasks by 16--50 points, with the largest gain on Eraser.
Pen reaches 90\%, providing the full-data reference below.

\paragraph{Generalization Transfer.}
All three experiments use the \textbf{Pen task}, fixing the manipulation objective while changing visual conditions.
\modelname{} reaches 72\% under Flash+Camera Offset and 68\% on the replaced Desk, each 10 points above $\pi_{0.5}$.
On Color, it reaches 64\%, versus 62\% for $\pi_0$.
These scores compare directly with the clean Pen reference.
Relative to that reference, \modelname{} retains 80.0\% under F+C and 75.6\% under Desk, compared with 70.5\% and 65.9\% for $\pi_{0.5}$, respectively.
This task-matched comparison measures transfer across changes in illumination, viewpoint, and tabletop appearance.

\paragraph{\blue{Reduced-Data Training.}}
This experiment also uses the \textbf{Pen task} on Agibot-G1, with \blue{400 of its 3,965 demonstrations}.
\modelname{} achieves \textbf{78\%}, versus 68\% for $\pi_0$, 66\% for $\pi_{0.5}$, and 30\% for GR00T-N1.7, exceeding the strongest baseline by 10 points.
It retains 86.7\% of its full-data Pen performance (78\% vs.\ 90\%); $\pi_{0.5}$ retains 75.0\% (66\% vs.\ 88\%).
Their gap increases from 2 to 12 points, supporting the value of motion-semantic representations with fewer demonstrations.

\begingroup
\setlength{\intextsep}{4pt}
\begin{table}[H]
\centering
\caption{
\textbf{Real-world evaluation on Agibot-G1.}
Success rates (\%) for eight standard desk tasks. All three generalization experiments---F+C (strobe lighting with camera offset), Desk (replaced surface), and Color (changed object colors)---and the \blue{reduced-data experiment} are conducted on the \textbf{Pen task}. \blue{All methods share 400 Pen demonstrations for reduced-data training}; Avg.\ is the arithmetic mean within each setting. Era.: eraser; CF: correction fluid; Chg.: charger; PC: pencil case; Sta.: stapler; Tr.: trash. Best results are \textbf{bold}.
}\label{tab:realworld}
\TableFont
\tabcolsep=2pt
\begingroup
\begingroup\TableFont
\tabcolsep=2pt
\renewcommand\arraystretch{1.0}
\noindent\begin{tabular*}{\linewidth}{@{\extracolsep{\fill}}l|cccccccc|c|ccc|c|c@{}}
\toprule
\multirow{2}{*}{\textbf{Method}}
& \multicolumn{8}{c|}{\textbf{Standard Tasks}}
& \multirow{2}{*}{\textbf{Avg.}}
& \multicolumn{3}{c|}{\textbf{Generalization}}
& \multirow{2}{*}{\textbf{Avg.}}
& \multirow{2}{*}{\makecell{\textbf{\blue{400}}\\\textbf{\blue{demos}}}}
\\
\cmidrule(lr){2-9} \cmidrule(lr){11-13}
& \rotatebox{0}{Pen}
& \rotatebox{0}{Era.}
& \rotatebox{0}{CF}
& \rotatebox{0}{Chg.}
& \rotatebox{0}{PC}
& \rotatebox{0}{Sta.}
& \rotatebox{0}{Book}
& \rotatebox{0}{Tr.}
& & \rotatebox{0}{F+C}
& \rotatebox{0}{Desk}
& \rotatebox{0}{Color}
& &
\\
\midrule

$\pi_{0}$
& 74 & 46 & 42 & 66 & 28 & 64 & 52 & 74 & 55.8
& 46 & 54 & 62 & 54.0 & 68 \\

GR00T-N1.7
& 38 & 52 & 30 & 64 & 24 & 74 & 80 & 82 & 55.5
& 18 & 24 & 28 & 23.3 & 30 \\

$\pi_{0.5}$
& 88 & 64 & \textbf{86} & 82 & 40 & 70 & \textbf{84} & 76 & 73.8
& 62 & 58 & \textbf{70} & 63.3 & 66 \\

\midrule
\rowcolor{black!5}
\textbf{\modelname{} (Ours)}
& \textbf{90} & \textbf{96} & 70 & \textbf{94} & \textbf{56} & \textbf{88} & 72 & \textbf{94} & \textbf{82.5}
& \textbf{72} & \textbf{68} & 64 & \textbf{68.0} & \textbf{78}

\\

\bottomrule
\end{tabular*}\par
\endgroup
\endgroup

\end{table}

\endgroup
\subsection{Ablation Studies}
\label{sec:exp:ablation}
Table~\ref{tab:ablation:component} shows the results of progressively adding components.
Starting from a bare backbone without future prediction or Motion-CoT adding future-frame prediction improves LIBERO by 2.3 points, indicating that the auxiliary future-frame objective provides useful visual grounding.
Replacing future-frame prediction with Motion-CoT alone yields a gain (0.9 point), indicating that distilled semantic tokens are a more effective information bottleneck than pixel-level generation.
Adding the \RDname{} branch with \CDEname{} supervision further boosts benchmarks (1.6 points), validating that explicit dynamics prediction regularizes motion-semantic tokens toward physically meaningful transitions.
Enabling causally-decoupled attention further improves LIBERO success by 0.8 percentage points, supporting the benefit of isolating motion-semantic queries from noised action tokens while allowing action tokens to attend to both the queries and visual context.
Finally, combining future-frame prediction with the full Motion-CoT pipeline yields the best results, demonstrating that the two complementary objectives-pixel-level visual grounding and semantic-level transition prediction-jointly achieve optimal performance.

\begin{figure}[htbp]
\centering
\begin{minipage}[t]{0.48\linewidth}
\vspace{0pt}
\captionsetup{type=figure}
\input{figures/expr2}
\end{minipage}\hfill
\begin{minipage}[t]{0.48\linewidth}
\vspace{0pt}
\setlength{\parskip}{0pt}
\captionsetup*{type=table}
\caption{\textbf{LIBERO component ablation.}
Fut.\ Pred: future-frame prediction; MotionCoT: queries; RD: Representational Deduction; CDE: Causal Dynamics Engine; Decoup.: decoupled attention. Checks/crosses: enabled/disabled; dashes: not applicable.
\blue{We further enhanced the performance on three libero-plus tasks to evaluate generalization.}
}
\label{tab:ablation:component}
\centering
\TableFont
\begingroup
\begingroup\TableFont
\tabcolsep=2pt
\renewcommand\arraystretch{1.0}
\noindent\begin{tabular}{ccccc|cccc}
\toprule
Fut.Pred & M.CoT & RD & CDE & Dec. & LIBERO & \blue{Cam}. & \blue{Rob.} & \blue{Lan.} \\
\midrule
\xmark & \xmark & -- & -- & -- & 91.3 & \blue{44.3} & \blue{45.6} & \blue{60.1} \\
\cmark & \xmark & -- & -- & -- & 93.6 & \blue{60.7} & \blue{49.8} & \blue{71.7} \\
\xmark & \cmark & \xmark & \xmark & \xmark & 94.5 & \blue{49.1} & \blue{47.5} & \blue{62.4} \\
\xmark & \cmark & \cmark & \cmark & \xmark & 96.1 & \blue{65.9} & \blue{54.3} & \blue{77.9} \\
\xmark & \cmark & \cmark & \cmark & \cmark & 96.9 & \blue{69.3} & \blue{62.6} & \blue{80.5} \\
\rowcolor{black!5}
\cmark & \cmark & \cmark & \cmark & \cmark & \textbf{97.9} & \blue{76.0} & \blue{70.2} & \blue{84.1} \\
\bottomrule
\end{tabular}\par
\endgroup
\endgroup

\end{minipage}
\end{figure}

\subsection{Representational Analysis}
\label{sec:exp:repr}

\paragraph{CDE Predictive Representation Quality}
The \CDEname{} predicts future VJEPA2-AC representations conditioned on current state and Motion-CoT tokens.
In Figure~\ref{fig:tsne}, we compare four RoboCasa-GR1 pick-and-place tasks with different source locations and target containers.
The visualization suggests that the CDE can capture patterns in the evolution of future latent representations.

\paragraph{PCA RGB pseudo-color visualization.}
We apply PCA to reduce the 1408-dimensional predicted tokens to RGB channels for a 16$\times$16 pseudo-color image.
Figure~\ref{fig:pca_rgb} shows that the CDE predictions produce coherent spatial structures that closely match the ground-truth future PCA colors, while differing from the current state.
The spatial coherence of the predicted pseudo-color images confirms that the CDE does not merely predict token-level averages but captures structured, region-aware state transitions.

\section{Conclusion and Limitation}\label{sec:conclusion}
\paragraph{Conclusion.} 
We present PILOT, which combines future-representation supervision of Motion-CoT with noise-isolated action decoding. Current-context queries guide trajectory refinement alongside visual features, while auxiliary prediction branches are omitted at inference. 
The evaluations examine manipulation success, robustness under selected shifts, and data-efficient adaptation on Agibot-G1.
These results support transition-oriented representation learning as a practical complement to action supervision.

\paragraph{Limitation.}
The current implementation uses a fixed number of motion queries and a fixed future-prediction interval. 
Adapting these choices to task duration may improve flexibility. 
Moreover, our visualizations are qualitative and do not assign a human-readable meaning to individual query slots. 
Developing lightweight diagnostics for these representations is a useful direction for future work.
\label{page:main-end}

\clearpage
\newpage
\section*{AI Use Statement}
Generative AI assisted manuscript preparation, including \LaTeX{} formatting, layout adaptation, citation and consistency checks, prose revision, and removal of repeated exposition.

\section*{Reproducibility Statement}
The model formulation and training objectives are specified in Section~\ref{sec:method}. The appendix provides training and inference pseudocode, architecture and optimization hyperparameters, evaluation protocols, dataset descriptions, and complete task-level RoboCasa-GR1 results (Section~\ref{sec:appendix:robocasa_results}).

\bibliographystyle{iclr2027_conference}
\bibliography{main}

\clearpage
\appendix
\begingroup
\color{black}
\raggedbottom
\arrayrulecolor{black}
\captionsetup{font={small,color=black},labelfont={bf,color=black}}
\renewcommand\arraystretch{1.2}

\section{\blue{Additional Analyses and Experiments}}
\label{sec:appendix:additional}

This section quantifies the contributions of Motion-CoT and world-model
context, analyzes predictive conditioning, clarifies evaluation variability
and reduced-data protocols, and reports additional real-world and
motion-based baseline comparisons.
The RoboCasa analyses below use separate trajectory and scene batches
from the 1200-episode released-checkpoint benchmark in
Section~\ref{sec:appendix:robocasa_results}; they are not a rerun of
that evaluation.

\subsection{\blue{Contributions of Motion-CoT and World-Model Context}}
\label{sec:appendix:contributions}

Using frozen PILOT weights trained on RoboCasa-GR1, we conduct three
analyses: attention allocation, separate blocking of the query and context
pathways, and replacement of query content while keeping the receiving
policy's inputs fixed.

\paragraph{Attention allocation.}
Motion queries (Q), world-model context (C), state (S), and actions (A)
contain 64/196/1/16 tokens, respectively. For each action row, we apply
softmax to the scaled query-key logits over all 277 source tokens and
sum the weights within each group. We average these group totals over
16 action rows, 16 heads, 16 layers, 20 flow steps, and 60 policy calls
per trajectory, then average trajectories equally. Q/token and C/token
divide the respective attention masses by 64 and 196.

\begin{table}[H]
\centering
\color{black}
\caption{\textbf{Action-token attention allocation on RoboCasa-GR1.}
Each task uses 10 trajectories. Group masses and per-token weights are
percentages; the final column is the Q-to-C per-token ratio.}
\label{tab:additional:attention}
\small
\setlength{\tabcolsep}{3pt}
\begin{tabular}{l|rrrr|rrr}
\toprule
\textbf{Task} & \makecell{Q mass\\(\%)} & \makecell{C mass\\(\%)}
& \makecell{S mass\\(\%)} & \makecell{A mass\\(\%)}
& \makecell{Q/token\\(\%)} & \makecell{C/token\\(\%)}
& \makecell{Q/C\\per token} \\
\midrule
CuttingboardToBasket & 39.44 & 53.71 & 1.14 & 5.71 & 0.616 & 0.274 & 2.248 \\
PlacematToPlate & 40.37 & 52.78 & 1.17 & 5.68 & 0.631 & 0.269 & 2.343 \\
PlateToBowl & 39.93 & 53.13 & 1.14 & 5.80 & 0.624 & 0.271 & 2.302 \\
CuttingboardToPan & 39.87 & 53.29 & 1.14 & 5.71 & 0.623 & 0.272 & 2.291 \\
TrayToCardboardbox & 39.77 & 53.34 & 1.13 & 5.76 & 0.621 & 0.272 & 2.283 \\
\bottomrule
\end{tabular}
\end{table}

As shown in Table~\ref{tab:additional:attention}, the 64 motion-query
tokens receive 39.44--40.37\% of total action attention, compared with
52.78--53.71\% for the 196 context tokens. Per-token weights are
0.616--0.631\% and 0.269--0.274\%, respectively: Motion-CoT receives
2.248--2.343 times the attention per token of context. These measurements
quantify attention allocation; the following interventions test whether
successful execution depends on the two pathways.

\paragraph{Separately blocking information pathways.}
The \textit{w/o query-KV} variant prevents action tokens from reading
motion-query keys/values while retaining world-model context.
The \textit{w/o direct WM-context} variant blocks direct reads of
world-model context while retaining motion-query input.
Both preserve raw proprioception and action self-attention.
We also block the corresponding reads by state tokens in every layer,
closing the indirect query/context-to-state-to-action route.

\begin{table}[H]
\centering
\color{black}
\caption{\textbf{Separate query and context blocking.}
Entries are successful rollouts out of 12. Each intervention is paired
with its original-policy control within its own scene batch.}
\label{tab:additional:blocking}
\small
\setlength{\tabcolsep}{4pt}
\begin{tabular}{l|cc|cc}
\toprule
\textbf{Task} & \makecell{Query-batch\\original}
& \makecell{W/o\\query-KV}
& \makecell{Context-batch\\original}
& \makecell{W/o direct\\WM-context} \\
\midrule
CuttingboardToPot & 8/12 & 1/12 & 11/12 & 0/12 \\
TrayToPlate & 10/12 & 0/12 & 8/12 & 0/12 \\
\bottomrule
\end{tabular}
\end{table}

Removing query access changes success from 66.7\% to 8.3\% on
CuttingboardToPot and from 83.3\% to 0\% on TrayToPlate.
Removing direct context access changes success from 91.7\% to 0\%
and from 66.7\% to 0\%, respectively
(Table~\ref{tab:additional:blocking}).
Visual-language context does not recover performance when query access
is removed; context-conditioned queries likewise do not recover performance
when direct context access is removed. Successful execution in the
trained policy depends on both reading pathways.

\paragraph{Replacing query content.}
Receiver and Donor are two inference passes through the same frozen
model. At each Receiver observation, both passes use the same image,
robot state, and noisy action tokens, but the Donor receives a modified
instruction. At every layer, we replace the Receiver's query-token
keys/values with the corresponding Donor keys/values. The Receiver retains
its original visual-language context, proprioception, and action noise.
An identical-input control, in which the Donor also receives the original
instruction, reproduces the original actions.

We compare two content interventions:
\begin{itemize}[leftmargin=*,nosep]
\item \textbf{Same action type, different target receptacle.}
Both instructions require picking and placing the same object, but
designate visually different receptacles in the same scene.
For example, the Receiver is instructed to place the eggplant in the pot,
whereas the Donor is instructed to place it in the bowl.
The image is unchanged; only the instruction-designated destination changes.
\item \textbf{Same object, different action type.}
The instructions refer to the same object but request incompatible
operations: for example, placing the eggplant in the pot versus cutting
the eggplant with a knife.
\end{itemize}

\begin{table}[H]
\centering
\color{black}
\caption{\textbf{Query-content replacement.}
Success is measured against the Receiver's original instruction using
12 paired scenes per task and condition.}
\label{tab:additional:query_swap}
\small
\setlength{\tabcolsep}{5pt}
\begin{tabular}{l|ccc}
\toprule
\textbf{Task} & Original
& \makecell{Same action type,\\different destination}
& \makecell{Same object,\\different action type} \\
\midrule
CuttingboardToPot & 8/12 & 6/12 & 0/12 \\
TrayToPlate & 10/12 & 10/12 & 1/12 \\
\bottomrule
\end{tabular}
\end{table}

Preserving the action type while changing the destination retains
substantial performance: success changes from 66.7\% to 50.0\% on
CuttingboardToPot and remains at 83.3\% on TrayToPlate.
Changing the action type while retaining the same object reduces success
to 0\% and 8.3\%, respectively (Table~\ref{tab:additional:query_swap}).
Thus, incompatible action content disrupts execution more strongly than
a conflicting destination, despite unchanged Receiver inputs.
Together with pathway blocking, these results support the use of
motion-specific query content alongside direct world-model context.
The destination intervention changes the instruction, not the image,
and is therefore not an isolated appearance perturbation.
\label{page:additional:contributions-end}

\subsection{\blue{Predictive Information Beyond the Current Scene and State}}
\label{sec:appendix:predictive}

We analyze both the dependence of the pretrained CDE on Motion-CoT and
the effect of training with different predictive inputs.

\paragraph{Frozen-model representation analysis on RoboCasa-GR1.}
For each task, we select 5 reference trajectories and 25 evaluation
trajectories. We sample four times $t\in\{0,120,360,600\}$ per
trajectory, forming current-future pairs $(I_t,I_{t+16})$.
This gives 20 reference pairs and 100 evaluation pairs per task.
The frozen encoder produces $z_t=E(I_t)$ and $z_{t+16}=E(I_{t+16})$,
and the frozen policy produces $m_t$ from the current observation,
instruction, and proprioceptive state $s_t$.
Writing CDE input projections implicitly, we compare:
\begin{itemize}[leftmargin=*,nosep]
\item \textbf{Original:} $G(z_t,s_t,m_t)$ uses the observation-specific
Motion-CoT.
\item \textbf{Zero query:} $G(z_t,s_t,0)$ retains current features and
robot state but replaces Motion-CoT with zeros.
\item \textbf{Reference mean:} $G(z_t,s_t,\bar m_{\mathrm{task}})$ uses
one fixed task representation, averaging the 20 reference Motion-CoTs
at each corresponding slot and feature dimension.
\item \textbf{Copy current:} $\hat z_{t+16}=z_t$ directly uses current
features as the future prediction.
\end{itemize}

\begin{table}[H]
\centering
\color{black}
\caption{\textbf{Future-feature SmoothL1 prediction error.}
The encoder, policy, and CDE are frozen; lower error is better.}
\label{tab:additional:prediction_error}
\small
\setlength{\tabcolsep}{4pt}
\begin{tabular}{l|cccc}
\toprule
\textbf{Task} & Original & Zero query & Reference mean & Copy current \\
\midrule
CuttingboardToPot
& $0.0154\pm0.0012$ & $0.1042\pm0.0075$
& $0.0261\pm0.0023$ & $0.2082\pm0.0161$ \\
TrayToPlate
& $0.0151\pm0.0015$ & $0.0953\pm0.0082$
& $0.0228\pm0.0026$ & $0.2079\pm0.0227$ \\
\bottomrule
\end{tabular}
\end{table}

The reference mean substantially improves over zero queries, suggesting
that a shared motion pattern supplies a useful task-level prior.
Observation-specific Motion-CoT further reduces error by 33.8--41.0\%
relative to the reference mean (Table~\ref{tab:additional:prediction_error}).
The shared pattern must therefore still be adapted to the current scene
and robot configuration. Zero queries outperform copying current
features because current features and robot state already provide some
cues about future evolution. Motion-CoT adds episode-specific transition
guidance beyond these cues and a fixed task prior.

\paragraph{Training with different predictive inputs.}
All variants retain query tokens as action-generation context.
They differ in whether the CDE reads those queries and whether
reconstruction gradients supervise them:
\begin{itemize}[leftmargin=*,nosep]
\item \textbf{Vanilla query token.}
Queries are trained only through action loss.
A separate CDE predicts
$\hat z_{t+\Delta}=G_\phi(z_t,s_t)$,
without reading or supervising the queries.
\item \textbf{Stop gradient.}
Both action generation and the CDE read the queries, but
$\hat z_{t+\Delta}=G_\phi(z_t,s_t,\operatorname{stopgrad}(m_t))$
blocks reconstruction gradients to the query-producing model,
where $m_t=F_\theta(\mathrm{current\ context},s_t)$.
Queries are trained only through action loss.
\item \textbf{Motion-CoT (PILOT).}
Both pathways read the queries, and action and CDE losses jointly
train them, with
$\hat z_{t+\Delta}=G_\phi(z_t,s_t,m_t)$.
\end{itemize}

\begin{table}[H]
\centering
\color{black}
\caption{\textbf{Effect of predictive inputs and query supervision.}
CDE loss is lower-is-better; LIBERO success is higher-is-better.
Loss summaries use the last 100 training steps and reflect
minibatch-to-minibatch fluctuations, not independent training runs.}
\label{tab:additional:predictive_training}
\small
\setlength{\tabcolsep}{5pt}
\begin{tabular}{l|ccc}
\toprule
\textbf{Metric} & \makecell{Vanilla\\query token}
& \makecell{Stop\\gradient} & \makecell{Motion-CoT\\(PILOT)} \\
\midrule
CDE reconstruction loss
& $0.0253\pm0.005$ & $0.0191\pm0.003$ & $0.0185\pm0.002$ \\
LIBERO success (\%) & 94.5 & 94.5 & \textbf{97.9} \\
\bottomrule
\end{tabular}
\end{table}

Action-trained queries reduce CDE reconstruction loss by 24.5\% relative
to current features and state alone. RD supervision further reduces loss
by 3.1\%, while increasing control success by 3.4 percentage points
(Table~\ref{tab:additional:predictive_training}).
Prediction error alone therefore does not capture Motion-CoT's control
value: joint predictive and action supervision makes the representation
useful for anticipating state changes and generating their realizing
trajectories. This supports learning high-level guidance and low-level
actions inside the same action model; it does not by itself establish
superiority over every alternative supervision location.
\label{page:additional:predictive-end}

\subsection{\blue{Evaluation Variability and Reduced-Data Protocols}}
\label{sec:appendix:variability}

\paragraph{Variability across evaluation seeds.}
Each ablation configuration is evaluated using seeds 907, 908, and 909.
The weights for each configuration come from a single training run with
seed 42. Table~\ref{tab:additional:seeded_ablation} reports per-seed
results and their mean and sample standard deviation.
The standard deviations reflect evaluation-seed variability, not
variability across independent training runs.

\begin{table}[H]
\centering
\color{black}
\caption{\textbf{LIBERO component ablations across evaluation seeds.}
Abbreviations follow Table~\ref{tab:ablation:component}.
Mean $\pm$ SD summarizes the three evaluation seeds for each
single-training-run checkpoint.}
\label{tab:additional:seeded_ablation}
\small
\setlength{\tabcolsep}{3pt}
\begin{tabular}{ccccc|rrrc}
\toprule
Fut.Pred. & M.CoT & RD & CDE & Dec.
& Seed 907 & Seed 908 & Seed 909 & Mean $\pm$ SD \\
\midrule
\xmark & \xmark & -- & -- & --
& 90.6 & 92.5 & 90.8 & $91.3\pm1.04$ \\
\cmark & \xmark & -- & -- & --
& 93.1 & 92.8 & 94.9 & $93.6\pm1.14$ \\
\xmark & \cmark & \xmark & \xmark & \xmark
& 95.9 & 93.9 & 93.7 & $94.5\pm1.22$ \\
\xmark & \cmark & \cmark & \cmark & \xmark
& 96.8 & 96.2 & 95.3 & $96.1\pm0.75$ \\
\xmark & \cmark & \cmark & \cmark & \cmark
& 95.2 & 97.6 & 97.9 & $96.9\pm1.48$ \\
\rowcolor{black!5}
\cmark & \cmark & \cmark & \cmark & \cmark
& 97.0 & 98.3 & 98.4 & $\mathbf{97.9}\pm0.78$ \\
\bottomrule
\end{tabular}
\end{table}

PILOT outperforms the bare and future-supervised baselines in every
evaluation seed, with average gains of 6.6 and 4.3 percentage points.
Adding RD/CDE to the query pathway raises mean success from 94.5\%
to 96.1\%; decoupled attention and future prediction further raise it
to 96.9\% and 97.9\%, respectively.

\paragraph{Pen generalization and reduced-data protocol.}
We randomly select 400 complete trajectories from the 3,965 Pen
demonstrations and fix this subset for PILOT and all compared baselines.
Both the action-DiT loss and CDE reconstruction loss use this subset
for reduced-data training. For each generalization and reduced-data
setting, PILOT and the baselines are trained from scratch rather than
continued from a full-data Pen-trained model. Each condition uses
50 rollouts, with success measured as the fraction completing the task.
We use \textit{reduced-data training}, rather than \textit{few-shot
learning}, to describe the 400-demonstration setting.

\paragraph{Extension beyond Pen.}
The Book upright-placement experiment in
Appendix~\ref{sec:appendix:book} and Table~\ref{tab:additional:book}
extends this evidence to grasping, orientation adjustment, and precise
placement into a rack. PILOT and $\pi_{0.5}$ share a fixed subset of
400 Book trajectories for reduced-data training.
PILOT improves over $\pi_{0.5}$ by 10/15/15 points under
lighting/camera, desk, and object-color changes, and by 25 points in
the reduced-data setting. The full protocol and results are reported
once in the next subsection.
\label{page:additional:variability-end}

\subsection{\blue{Book Generalization and Reduced-Data Evaluation}}
\label{sec:appendix:book}

Book upright placement requires grasping, orientation adjustment, and
precise placement into a storage rack. For reduced-data training,
PILOT and $\pi_{0.5}$ use the same fixed subset of 400 Book trajectories.
The Standard column in Table~\ref{tab:additional:book} retains the
full-data results from 50 rollouts.
The next three columns evaluate lighting and camera changes (F+C),
a changed desk surface/background (Desk), and changed object color
(Color), using 20 rollouts per condition.
The final column evaluates the reduced-data models under standard
visual conditions, also using 20 rollouts.

\begin{table}[H]
\centering
\color{black}
\caption{\textbf{Book upright-placement success rates (\%).}
Standard uses 50 rollouts; every other condition uses 20.
Reduced-data models use 400 demonstrations.
Best results in each column are bold.}
\label{tab:additional:book}
\small
\setlength{\tabcolsep}{6pt}
\begin{tabular}{l|rrrr|r}
\toprule
\textbf{Method} & Standard & F+C & Desk & Color
& \makecell{Reduced-data\\(400 demos)} \\
\midrule
$\pi_{0.5}$ & \textbf{84} & 25 & 35 & 40 & 20 \\
\rowcolor{black!5}
\textbf{PILOT} & 72 & \textbf{35} & \textbf{50}
& \textbf{55} & \textbf{45} \\
\bottomrule
\end{tabular}
\end{table}

PILOT improves over $\pi_{0.5}$ by 10/15/15 percentage points under
F+C/Desk/Color and by 25 points with reduced data.
It retains 62.5\% of its full-data success with 400 demonstrations,
compared with 23.8\% for $\pi_{0.5}$.
Despite $\pi_{0.5}$'s higher full-data result, PILOT performs better
in every tested generalization and reduced-data condition.
These results extend the robustness and data-efficiency evidence beyond
Pen to orientation-sensitive placement. The term \textit{reduced-data}
and the explicit demonstration count delimit the scope of this evidence.
\label{page:additional:book-end}

\subsection{\blue{Comparison with Motion-Based VLA Methods}}
\label{sec:appendix:motion_baselines}

Table~\ref{tab:additional:motion_baselines} compares PILOT with the
reported standard-LIBERO results of HiF-VLA~\citep{lin2025hifvla}
and CoWVLA~\citep{yang2026cowvla}.
The baseline results come from Table~4 of HiF-VLA and Table~1 of
CoWVLA, respectively, and retain their original evaluation protocols.

\begin{table}[H]
\centering
\color{black}
\caption{\textbf{Standard-LIBERO comparison with motion-based methods.}
Success rates (\%); \colorbox{black!25}{best} and
\colorbox{black!10}{second-best} values are shaded per column.}
\label{tab:additional:motion_baselines}
\small
\setlength{\tabcolsep}{7pt}
\begin{tabular}{l|rrrr|r}
\toprule
\textbf{Method} & Spatial & Object & Goal & Long & Average \\
\midrule
HiF-VLA & \best{98.8} & \best{99.4} & \second{97.4}
& \second{96.4} & \best{98.0} \\
CoWVLA & \second{97.2} & 97.8 & 94.6 & 92.8 & 95.6 \\
PILOT & 97.1 & \second{99.2} & \best{98.1}
& \best{97.2} & \second{97.9} \\
\bottomrule
\end{tabular}
\end{table}

PILOT achieves comparable overall success to HiF-VLA
(97.9\% versus 98.0\%), with higher Goal and Long success by
0.7 and 0.8 percentage points.
Compared with CoWVLA, PILOT improves the average by 2.3 points,
with gains of 1.4/3.5/4.4 points on Object/Goal/Long.
Its strongest relative results are on goal-conditioned and
long-horizon manipulation.

The methods also differ in how motion guidance is learned.
HiF-VLA integrates motion-vector representations with action learning,
whereas CoWVLA learns latent motion from video representations
factorized into structure and motion.
PILOT learns queries inside the flow-based action model and supervises
them as conditions for current-to-future representation prediction.
The same action model thus learns predictive motion guidance and
fine-grained trajectory generation, while asymmetric attention keeps
the guidance independent of action-sampling noise.
\label{page:additional:motion_baselines-end}

\par
\endgroup
\clearpage

\section{Extended Related Work}
\label{sec:appendix:related}

\begin{figure}[H]
\centering
\centerline{\includegraphics[width=\linewidth]{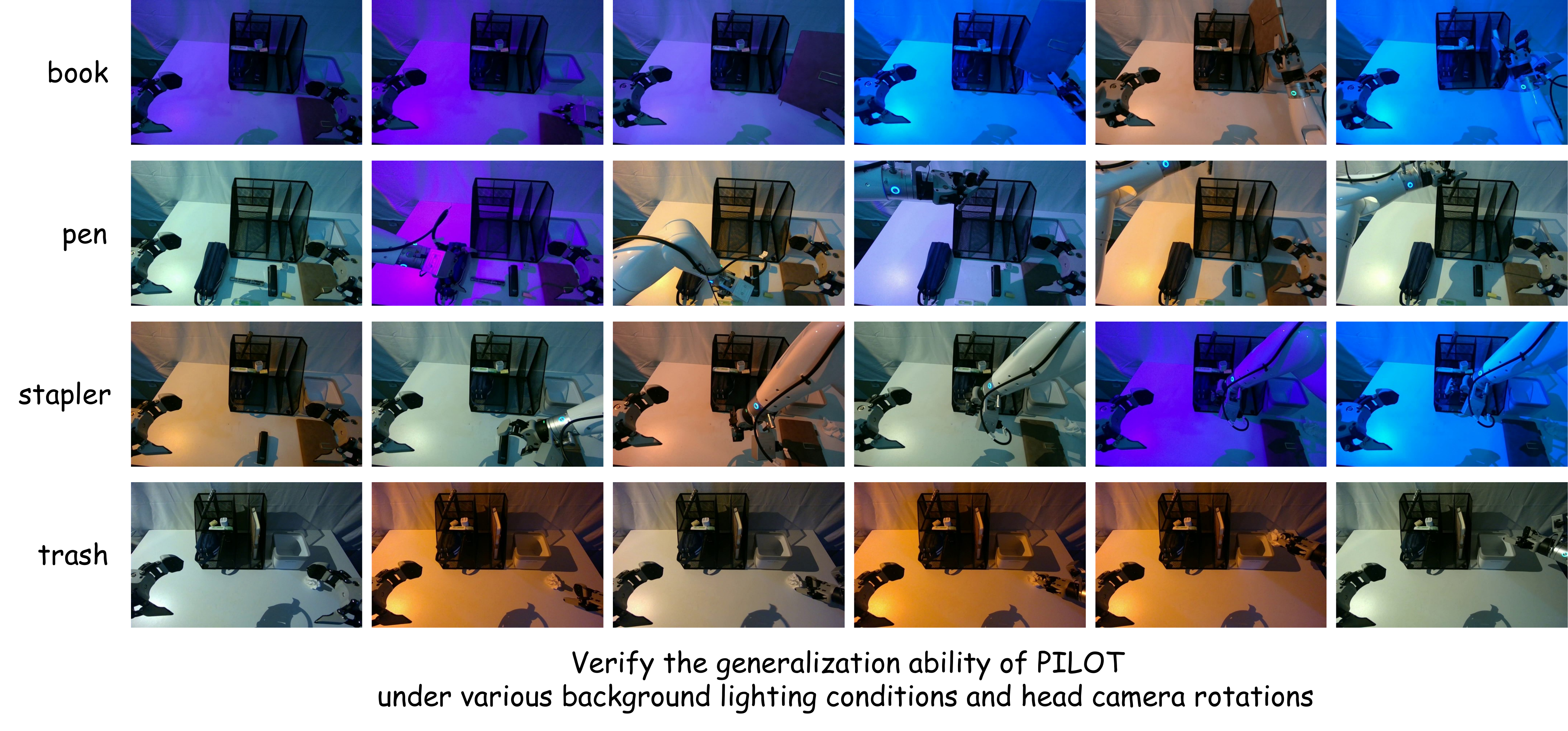}}
\caption{
Qualitative demonstrations under the \textit{Flash + Camera Offset} setting, with strobe-light interference and a shifted head-mounted camera. The examples include multiple tasks; the quantitative generalization results in Table~\ref{tab:realworld} are evaluated on the Pen task only.
}\label{fig:append3}
\end{figure}

This section expands the related-work discussion in the main text.

\subsection{Vision-Language-Action Models}
\label{sec:appendix:related:vla}

Vision-Language-Action (VLA) models~\citep{brohan2022rt1,kim2024openvla,black2024pi0,intelligence2025pi05,bjorck2025gr00tn1} combine visual observations and language instructions with action prediction. Many recent designs adapt pretrained vision-language models (VLMs) using robot demonstrations.
RT-1~\citep{brohan2022rt1} predicts discretized actions with a robot-trained transformer, while RT-2~\citep{brohan2023rt2} adapts a pretrained vision-language model to predict action tokens.
This paradigm has been extended through several architectural innovations: parallel decoding for improved efficiency~\citep{kim2025openvlaoft}, flow-matching action experts for continuous action spaces~\citep{black2024pi0,intelligence2025pi05}, diffusion-based action prediction~\citep{liu2025rdt}, and compact open-source VLMs for broader accessibility.

Recent efforts have scaled VLAs to larger backbones and more diverse embodiments. GR00T~\citep{bjorck2025gr00tn1} and the $\pi$ family~\citep{black2024pi0,intelligence2025pi05} combine vision-language backbones with continuous action generation and heterogeneous robot data. UniVLA~\citep{bu2025univla} learns task-centric latent actions, whereas CogVLA~\citep{li2026cogvla} uses instruction-driven routing and sparsification. SpatialVLA~\citep{qu2025spatialvla} introduces Ego3D position encoding and adaptive discretized action grids. ROCKET~\citep{sun2026rocketvla} aligns multiple VLA layers with a geometric foundation model through a shared residual-stream projector. GeoAware-VLA~\citep{abouzeid2025geoaware} uses a frozen geometric vision encoder and a trainable projection into the policy decoder.

\subsection{World Action Models}
\label{sec:appendix:related:wam}

World Action Models (WAMs)~\citep{cen2025worldvla,li2026lingbotva} leverage world modeling, predicting future visual states, to support downstream action prediction. We follow~\citep{cen2025worldvla} to use this terminology, as these models unify world state evolution modeling and action generation within a single framework.

WAMs differ in how prediction interacts with control. In \emph{predict-then-act} designs, predicted visual states supply inputs for planning or action decoding. Mimic-Video~\citep{pai2025mimic} conditions flow-based control on intermediate features of its video generator, avoiding the need to reconstruct a complete video before producing actions.
ImageWAM~\citep{zhang2026imagewam} studies image editing as an alternative to video generation for learning action models.

Other designs jointly model future states and actions. Motus~\citep{bi2026motus} integrates video and action modeling through a mixture-of-transformers architecture. LDA, short for Latent Dynamics Action~\citep{lyu2026lda}, models dynamics and actions in a DINO-based representation space using a multimodal diffusion transformer. Cosmos Policy~\citep{kim2026cosmospolicy} incorporates actions and future values into a video model's diffusion process. FastWAM~\citep{yuan2026fastwam} compares training-time video co-modeling with test-time future imagination and reports that video co-training accounts for the main gains in its experiments. DreamZero~\citep{ye2026dreamzero} jointly generates video and actions with system-level deployment optimizations; its Flash variant additionally decouples video and action noise schedules.

PILOT complements its auxiliary future-frame objective with RD supervision in the representation space of a frozen VJEPA2-AC encoder. During training, the CDE predicts a future representation from current features, motion-semantic queries, and robot state. The action-only inference path retains direct visual context together with Motion-CoT and does not require the auxiliary prediction branches. This combination couples transition-oriented supervision with a dedicated, noise-isolated query pathway rather than requiring future images as an intermediate input to the policy.

\subsection{Joint Embedding Predictive Architectures}
\label{sec:appendix:related:jepa}

Joint Embedding Predictive Architectures (JEPA)~\citep{assran2025vjepa2,mur2026vjepa21} predict target features in a learned representation space rather than reconstructing pixels. This changes the prediction target, but does not by itself guarantee invariance to texture or the recovery of physical state variables.

VJEPA2-AC~\citep{assran2025vjepa2} couples an observation encoder with an action-conditioned predictor. PILOT uses the former to encode current and future observations and initializes the trainable CDE from the latter. LeWorld~\citep{maes2026leworldmodel} applies JEPA-style prediction to world modeling for planning. A comparative study evaluates reconstruction-oriented and semantic encoders, including VJEPA2.1, for robotic world models~\citep{jha2026reconstruction}. WoG instead learns a compact action-conditioning space from pretrained visual encoders and, in its second stage, predicts those conditions from current inputs~\citep{su2026wog}.
VLA-JEPA~\citep{sun2026vlajepa} learns time-indexed latent action tokens from current inputs through future-state prediction, retaining this objective alongside action loss during robot fine-tuning. PILOT instead learns unordered query slots within its flow-matching action head and jointly optimizes action, future-video, and future-representation objectives.

\subsection{Latent Action Models}
\label{sec:appendix:related:lam}

Latent action learning derives compact transition representations from video. LAPA~\citep{ye2025lapa} learns discrete latent actions without action labels and uses them for policy pretraining. \mbox{UniVLA~\citep{bu2025univla}} learns task-centric discrete actions and adapts them to robot controls. These are specific designs, rather than a single architecture shared by all latent-action methods. CoMo instead learns continuous latent motion using early temporal differences and temporal contrastive learning~\citep{yang2026como}.

Being-H0.7~\citep{luo2026beingh07} aligns latent-query hidden states in a current-input prior branch with a training-only, future-conditioned posterior branch. WoG learns action conditions using multi-encoder queries~\citep{su2026wog}. CoWVLA factorizes video into structure and motion latents, learns a continuous motion chain from an instruction and initial frame, and aligns that representation with action decoding~\citep{yang2026cowvla}. JALA learns transition-aware predictive action embeddings aligned with inverse dynamics and real actions, bypassing full visual reconstruction~\citep{luo2026jala}.

PILOT combines three choices: frozen VJEPA future representations as supervision targets, motion-semantic queries derived from current observation context and state, and trained through future-representation supervision, and asymmetric attention within the Action-Perceiver. Future frames supervise the CDE and auxiliary video branch during training. The deployed policy uses current inputs, with actions attending to both queries and visual context. Thus, the comparison concerns the target representation, query construction, and decoder information flow.

\begin{figure}[t]
\centering
\centerline{\includegraphics[width=\linewidth]{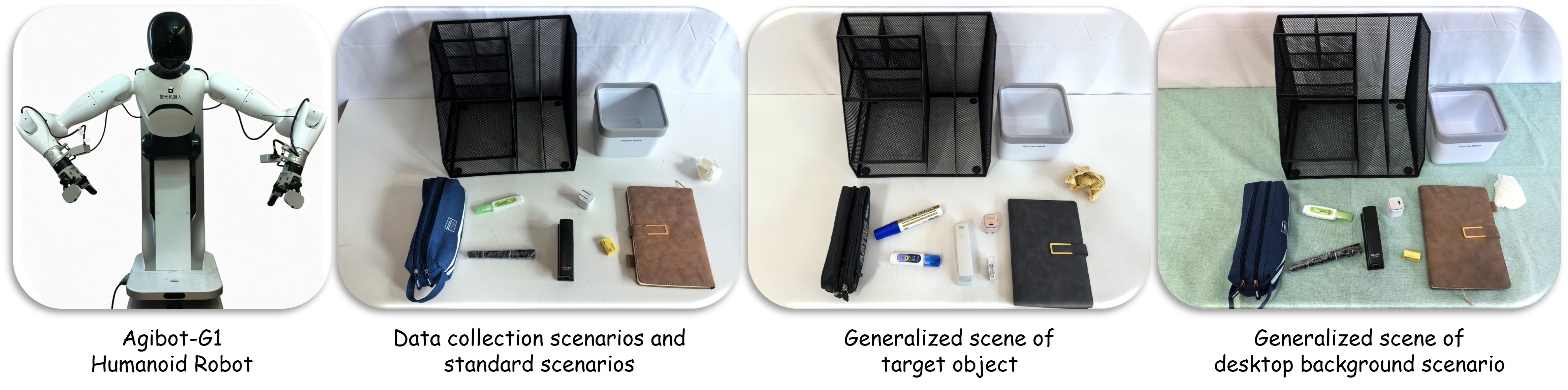}}
\caption{
Overview of the real-world experimental setup. From left to right: the Agibot-G1 dual-arm humanoid robot, the standard tabletop scene, a scene with changed target objects, and a scene with a replaced desk surface. These scene illustrations are distinct from the Flash + Camera Offset demonstrations in Figure~\ref{fig:append3}.
}\label{fig:append1}
\end{figure}

\subsection{Flow-Matching for Action Generation}
\label{sec:appendix:related:flow}

Flow matching learns a continuous-time vector field for transforming a prior distribution into a data distribution~\citep{tong2023otcfm}. In robotic action generation, $\pi_0$~\citep{black2024pi0} uses a flow-matching action expert to generate continuous action chunks, and $\pi_{0.5}$~\citep{intelligence2025pi05} extends the training recipe with heterogeneous data and hierarchical prediction. By contrast, OpenVLA-OFT~\citep{kim2025openvlaoft} uses Optimized Fine-Tuning: parallel decoding, action chunking, continuous action representations, and an L1 regression objective.

The action objective in Eq.~\ref{eq:loss-action} regresses the velocity along a linear interpolation between noise and a target action chunk. The decoder combines visual context and Motion-CoT to produce an action chunk without requiring auxiliary future prediction.

\section{Implementation Details}
\label{sec:appendix}

\subsection{Architecture and Information Flow}
\label{sec:appendix:arch_details}
\label{sec:appendix:hyper}

We follow the notation of Eq.~\ref{eq:factor}: $H$ is the action
horizon, and $d_s$ and $d_a$ are the state and action dimensions.
RD is designed to learn predictive, action-relevant state-transition
representations, with asymmetric attention enforcing directed information flow.

PILOT uses a trainable Cosmos-Predict2.5-2B transformer
\citep{nvidia2025cosmospredict25} in two modes that share weights.
The understanding mode extracts the last-layer context from one
conditional forward pass on the current observation and instruction.
The auxiliary generation mode predicts future video latents during
training. The accompanying text encoder and VAE are frozen.
At a world-model input resolution of $224\times224$, the understanding
path returns $N_c=196$ context tokens of width $d_c=2048$.
The implementation samples the VAE latent distribution rather than
using its mean. Query invariance to action noise is conditional on the
resulting context, not a claim that repeated image encoding is deterministic.

\paragraph{Action-Perceiver.}
The action head has 16 Perceiver blocks \citep{jaegle2021perceiver}
and 16 attention heads.
The Action-Perceiver uses an internal width of $d_h=768$ and produces
Motion-CoT outputs of width $d=1024$.
One state token, $K=64$ learned query tokens, and $H$ noised action
tokens form the latent sequence. The context is projected from 2048
to 768 dimensions before attention.
State and query tokens attend to their own group and the visual context;
action tokens additionally attend to the noised action group.
Time conditioning is restricted to action tokens. Thus, action noise
and flow time do not enter the query pathway.
The final query outputs constitute Motion-CoT, with no prescribed
temporal ordering of the $K$ slots. A
$1024\to1024\to d_a$ MLP decodes the action-token outputs into velocities.

\paragraph{Representational Deduction.}
The frozen VJEPA2-AC ViT-G encoder \citep{assran2025vjepa2} supplies
$P=256$ patch features of width $d_r=1408$.
Current and future images are processed separately, each duplicated into
a two-frame clip and resized/cropped to $256\times256$ using the encoder's
preprocessing. Layer-normalized future features are supervision targets,
not inputs to the policy or predictor.

The CDE is initialized from the 24-layer VJEPA2-AC predictor, with
internal width $d_p=1024$. Current patch features are projected
from 1408 to 1024 dimensions. Current patch features are projected from 1408 to 1024 dimensions. Motion and state tokens are first projected to the encoder feature dimensionality, $d_r=1408$, and subsequently mapped to the predictor width, $d_p=1024$. We denote the resulting predictor-space conditioning tokens by $\tilde{\bm m}$ and $\tilde{\bm s}$:
\begin{equation}
  \tilde{\bm m}\in\mathbb{R}^{K\times d_p},
  \qquad \tilde{\bm s}\in\mathbb{R}^{1\times d_p}.
  \label{eq:app:cde-conditioning}
\end{equation}
These conditions are prepended to the current patch tokens, giving
$K+1+P=321$ tokens. The predictor retains the patch outputs and maps
them back to 1408 dimensions. The original predictor's action, state,
and extrinsics encoders are bypassed and frozen.
The conditioning width $d_p$ must not be confused with the encoder
output width $d_r$.

\begin{table}[t]
\centering
\caption{Architecture of the supplied implementation. Input/internal
and output widths are distinguished explicitly.}
\label{tab:hyper}
\small
\begin{tabular}{ll}
\toprule
Component & Configuration \\
\midrule
World-model transformer & 28 layers; 196 context tokens, width 2048 \\
Action-Perceiver & 16 layers, 16 heads; internal width 768 \\
Motion-CoT & 64 query slots; output width 1024 \\
Action head FFN & Expansion factor 4, squared ReLU \\
VJEPA encoder features & $256\times1408$ patch grid \\
CDE & 24 layers; internal width 1024; 321 input tokens \\
\bottomrule
\end{tabular}
\end{table}

\subsection{Parameter Breakdown}
\label{sec:appendix:parameters}
For PILOT, the total parameter count is approximately 11.956B, of which 2.525B are trainable.
Tables~\ref{tab:libero} and~\ref{tab:libero_plus} report the trainable count,
rounded to 2.5B and marked with $^\dagger$.
Table~\ref{tab:parameter_breakdown} lists the component counts and training status.
The shared Cosmos transformer is counted once across its understanding
and generation modes.

\begin{table}[t]
\centering
\caption{PILOT parameter breakdown. Counts are approximate and reported
in billions (B).}
\label{tab:parameter_breakdown}
\small
\begin{tabular}{lrl}
\toprule
Module & Parameters (B) & Training status \\
\midrule
Cosmos-Predict2.5 DiT & 2.059 & Trainable \\
Cosmos-Reason1 / Qwen2.5-VL encoder & 8.292 & Frozen \\
Wan VAE & 0.127 & Frozen \\
V-JEPA2 ViT-g encoder & 1.012 & Frozen \\
V-JEPA2-AC predictor & 0.305 & Primarily trainable \\
Time-aware Perceiver action head & 0.160 & Trainable \\
State and feature projectors & 0.0011 & Trainable \\
\midrule
\textbf{Total} & \textbf{11.956} & \\
Trainable & 2.525 & \\
Frozen & 9.431 & \\
\bottomrule
\end{tabular}
\end{table}

\subsection{ActionEncoder and Token Initialization}
\label{sec:appendix:action_enc}
The following equations specialize to the RoboCasa configuration;
benchmark-specific horizons and interface dimensions are listed in
Table~\ref{tab:train_hyp}. We omit the batch dimension and
write $\mathrm{Linear}_{u\to v}$ for an affine map. The action horizon
is $H=16$. Given $\bm a^\tau\in\mathbb R^{H\times d_a}$,
the action time is discretized as
$q_\tau=\operatorname{trunc}(1000\tau)$, where truncation is toward zero,
matching integer conversion in the implementation:
\begin{align}
 \bm e_a &= \mathrm{Linear}_{d_a\to d_h}(\bm a^\tau),\\
 \omega_j &= 10000^{-j/(d_h/2)},\quad j=0,\ldots,d_h/2-1,\\
 \bm e_\tau &= [\,\sin(q_\tau\bm\omega);
                         \cos(q_\tau\bm\omega)\,],\\
 \bm A_{\rm enc} &=
 \mathrm{Linear}_{d_h\to d_h}
 \left(\mathrm{SiLU}\left(
 \mathrm{Linear}_{2d_h\to d_h}
 ([\,\bm e_a;\operatorname{repeat}_H(\bm e_\tau)\,])
 \right)\right).
 \label{eq:app:action-encoder}
\end{align}
The concatenation in Eq.~\ref{eq:app:action-encoder} is along the
feature dimension; sequence concatenations below are along the token
dimension. Learned horizon-position embeddings
$\bm p_i\in\mathbb R^{768}$ distinguish the $H$ action positions.
The Perceiver has a separate time embedding:
\begin{align}
 \bm h_\tau &=
 \mathrm{Linear}_{768\to768}\left(\mathrm{SiLU}\left(
 \mathrm{Linear}_{320\to768}(\operatorname{SinCos}_{320}(q_\tau))
 \right)\right),\\
 \bm b_\tau &= \mathrm{Linear}_{768\to768}
                         (\mathrm{SiLU}(\bm h_\tau)),\\
 \bm A^{(0)}_i &= (\bm A_{\rm enc})_i+\bm p_i+\bm b_\tau,\\
 \bm S &= \mathrm{Linear}_{1024\to768}
               \left(\mathrm{ReLU}(\mathrm{Linear}_{64\to1024}(s_t))\right),
 \qquad \bm Z^{(0)}=[\,\bm S;\bm Q\,].
 \label{eq:app:token-initialization}
\end{align}
Here $\operatorname{SinCos}_{320}$ uses the timestep encoder's
cosine-then-sine ordering, and $\bm Q\in\mathbb R^{64\times768}$ is
learned. Only the action tokens receive the horizon embedding and
time bias. The context is separately projected as
$\bm C=\mathrm{Linear}_{2048\to768}(\bm c)$.

\subsection{Layer-Level Action-Perceiver}
\label{sec:appendix:perceiver}
For layer $l=0,\ldots,15$, $\bm Z^{(l)}$ contains the state and query
tokens, and $\bm A^{(l)}$ contains action tokens. With layer-specific
parameters, adaptive normalization is
\begin{equation}
 \operatorname{AdaLN}(\bm X,\bm h_\tau)
 = (1+\bm\gamma_\tau)\odot\operatorname{LN}_0(\bm X)+\bm\beta_\tau,
 \quad
 [\,\bm\beta_\tau;\bm\gamma_\tau\,]
 =\mathrm{Linear}_{768\to1536}(\mathrm{SiLU}(\bm h_\tau)).
 \label{eq:app:adaln}
\end{equation}
$\operatorname{LN}_0$ has no learned affine parameters.
The modulation map is initialized to zero, so the multiplier is
$1+\bm\gamma_\tau$, not $\bm\gamma_\tau$ alone.
The attention and feed-forward sublayers have separate normalization
parameters. One complete block computes
\begin{align}
 \bar{\bm Z} &= \operatorname{LN}_{l,z}(\bm Z^{(l)}),&
 \bar{\bm A} &= \operatorname{AdaLN}_{l,a}(\bm A^{(l)},\bm h_\tau),&
 \bar{\bm C} &= \operatorname{LN}_{l,c}(\bm C),\\
 \bm U_Z &= \bm Z^{(l)}
       +\operatorname{MHA}_l(\bar{\bm Z},[\,\bar{\bm Z};\bar{\bm C}\,]),\\
 \bm U_A &= \bm A^{(l)}
       +\operatorname{MHA}_l(\bar{\bm A},
                  [\,\bar{\bm Z};\bar{\bm A};\bar{\bm C}\,]),\\
 \bm Z^{(l+1)} &= \bm U_Z+
               \operatorname{FFN}_l(\operatorname{LN}_{l,zf}(\bm U_Z)),\\
 \bm A^{(l+1)} &= \bm U_A+
               \operatorname{FFN}_l(
                     \operatorname{AdaLN}_{l,af}(\bm U_A,\bm h_\tau)).
 \label{eq:app:perceiver-block}
\end{align}
Both attention calls use the same module within a layer and the
\emph{pre-update} normalized tokens. In particular, $\bm U_Z$ is not
used as the key/value input for the action attention in that layer.
There are 16 heads of width 48. The shared feed-forward map is
\begin{equation}
 \operatorname{FFN}_l(\bm X)=
 \mathrm{Linear}_{3072\to768}
 \left(\operatorname{ReLU}(
        \mathrm{Linear}_{768\to3072}(\bm X))^2\right).
\end{equation}
The non-action and action branches have 261 and 277 key/value tokens,
respectively. Since the former sees neither $\bm A^{(l)}$ nor
$\bm h_\tau$, its outputs are invariant to action noise and time
conditional on fixed context and state.
This statement assumes the state token is present.
The final outputs are
\begin{align}
 \bm O &= \operatorname{LN}\left(
              \mathrm{Linear}_{768\to1024}
                     ([\,\bm Z^{(16)};\bm A^{(16)}\,])\right),\\
 \bm m &= \bm O_{\rm query},\qquad
 \hat{\bm v}=
 \mathrm{Linear}_{1024\to d_a}
       \left(\operatorname{ReLU}(
                    \mathrm{Linear}_{1024\to1024}(\bm O_{\rm action}))\right).
\end{align}
\subsection{Layer-Level Causal Dynamics Engine}
\label{sec:appendix:evolver}
Let $\bm r_t\in\mathbb R^{256\times1408}$ denote the layer-normalized
current-image representation. The CDE input is
\begin{align}
 \bm X_{\rm patch} &= \mathrm{Linear}_{1408\to1024}(\bm r_t),\\
 \tilde{\mathbf m}
&=\mathrm{Linear}_{1024\to1408}(\mathbf m),\\
\tilde{\mathbf s}
&=\mathrm{Linear}_{64\to1408}(\mathbf s_t),\\
\bar{\mathbf m}
&=P_m(\tilde{\mathbf m}),\\
\bar{\mathbf s}
&=P_s(\tilde{\mathbf s}),\\
 \bm X^{(0)} &= [\,\tilde{\bm m};\tilde{\bm s};\bm X_{\rm patch}\,]
                      \in\mathbb R^{321\times1024}.
 \label{eq:app:cde-input}
\end{align}
For each of the 24 predictor blocks,
\begin{align}
 \bm Y^{(l)} &= \bm X^{(l)}
       +\operatorname{Attn}^{\rm rope}_l(
                    \operatorname{LN}_{l,1}(\bm X^{(l)});\bm B),\\
 \bm X^{(l+1)} &= \bm Y^{(l)}
       +\mathrm{Linear}_{4096\to1024}\left(
          \operatorname{GELU}\left(
          \mathrm{Linear}_{1024\to4096}
                   (\operatorname{LN}_{l,2}(\bm Y^{(l)}))\right)\right).
 \label{eq:app:cde-block}
\end{align}
For each attention head, the normalized inputs are projected to
$\bm q,\bm k,\bm v$ of width 64. The attention output before head
concatenation and output projection is
\begin{equation}
 \operatorname{softmax}\left(
       \frac{\mathcal R(\bm q)\mathcal R(\bm k)^\top}{\sqrt{64}}
       +\bm B\right)\bm v.
 \label{eq:app:cde-attention}
\end{equation}
$\mathcal R$ denotes the inherited rotary implementation.
The 65 conditioning tokens use temporal position zero; patch tokens
use temporal position zero and their $16\times16$ spatial coordinates.
The predictor has 16 heads and no dropout in the supplied construction.
After the last block, the conditioning tokens are removed:
\begin{equation}
 \hat{\bm r}_{t+\Delta}
 =\mathrm{Linear}_{1024\to1408}
     \left(\operatorname{LN}
                 (\bm X^{(24)}_{\rm patch})\right)
       \in\mathbb R^{256\times1408}.
 \label{eq:app:cde-output}
\end{equation}

\paragraph{Attention-mask strategy.}
The supplied adaptation slices the pretrained frame-block mask rather
than rebuilding it for 65 conditioning tokens. Its original block size
is $256+2=258$. For the 321-token input, the additive mask actually used
in Eq.~\ref{eq:app:cde-attention}, with zero-based indices, is
\begin{equation}
 B_{ij}=
 \begin{cases}
  -\infty,&0\le i<258,\quad258\le j<321,\\
  0,&\text{otherwise}.
 \end{cases}
 \label{eq:app:cde-legacy-mask}
\end{equation}
This is not a
within-frame autoregressive mask or an unrestricted single-frame mask.
The reported model uses this attention-mask strategy.
No future-image features enter the CDE inputs.

\subsection{Training Objectives}
\label{sec:appendix:train}
\label{sec:appendix:future_objective}
\label{sec:appendix:action_dit}

The codebase builds on StarVLA \citep{ye2026starvla}.
Each training example provides current and future observations,
an instruction, robot state, and a target action chunk with its dimension
mask. The three objectives are optimized jointly:
\begin{equation}
  \mathcal L=\mathcal L_{\mathrm{act}}
  +\lambda_{\mathrm{wm}}\mathcal L_{\mathrm{wm}}
  +\lambda_{\mathrm{RD}}\mathcal L_{\mathrm{RD}}.
\end{equation}
For the released step-340000 RoboCasa checkpoint,
$\lambda_{\mathrm{wm}}=1$ and $\lambda_{\mathrm{RD}}=0.1$,
as recorded in its archived configuration.
The world-model transformer, action head, CDE, and conditioning
projectors are trainable; the VJEPA encoder, text encoder, and VAE
remain frozen.

\paragraph{Action flow matching.}
\label{sec:appendix:time}
Training uses a straight probability path \citep{tong2023otcfm},
$\bm a^\tau=(1-\tau)\bm\epsilon+\tau\bm a$, with Gaussian
$\bm\epsilon$ and velocity target $\bm a-\bm\epsilon$.
The squared error is averaged over the entries selected by the action
dimension mask. This mask excludes unused dimensions, not temporal
padding at episode boundaries.
The action time is sampled by
\begin{equation}
  b\sim\mathrm{Beta}(1.5,1.0),\qquad
  \tau=\operatorname{clip}\left(\frac{0.999-b}{0.999},\,0,\,1\right).
  \label{eq:app:beta-time}
\end{equation}
The clipped transformation ensures that the action time lies in $[0,1]$.
Each example is repeated $R=4$ times with independent action noise;
query outputs are averaged over replicas for RD supervision.
The repeats do not multiply the number of independent demonstration
examples in a batch.

\paragraph{Auxiliary future-latent objective.}
\label{sec:appendix:change_mask}
The VAE encodes a short clip consisting of the current image followed
by four repetitions of the future target. This gives one current and
one future latent group for the supplied single-image setting.
Current latents remain clean, and only the future group is noised.
With $u\sim\mathcal N(0,1)$, the world-model noise level is
\begin{equation}
  \tilde\tau'=\operatorname{sigmoid}(u),\qquad
  \tau'=\frac{5\tilde\tau'}{1+4\tilde\tau'}.
  \label{eq:app:logitnormal}
\end{equation}
The target interpolation is $(1-\tau')\bm z^*+\tau'\bm\epsilon'$,
whose velocity target is $\bm\epsilon'-\bm z^*$.
The auxiliary future-latent objective is:
\begin{equation}
  \mathcal L_{\mathrm{wm}}=
  \frac{\sum_j M_j
  \left(\hat v_j-(\epsilon'_j-z_j^*)\right)^2}{\sum_j M_j},
  \label{eq:app:wm-change}
\end{equation}
where $M$ selects future temporal positions.

\paragraph{Future-representation objective.}
The CDE predicts the layer-normalized future VJEPA features from
current features, Motion-CoT, and state. RD uses mean SmoothL1 loss
with the default transition parameter of one.
Future observations are used only in the auxiliary training branches.

\begin{algorithm}[t]
\caption{PILOT training}\label{alg:train}
\begin{algorithmic}[1]
\Require Current observation $o_t$, future observation $o_{t+\Delta}$,
instruction $\ell$, state $s_t$, actions $\bm a_t$, dimension mask $M$
\State $\bm c\gets\mathrm{WM}^{\mathrm{enc}}_\theta(o_t,\ell)$
\State Repeat current context, state, actions, and mask $R=4$ times
\State Sample independent action noise and clipped transformed Beta times
\State Compute action velocities, masked $\mathcal L_{\mathrm{act}}$,
and the average query output $\bm m$ using the Action-Perceiver
\State Compute future-only $\mathcal L_{\mathrm{wm}}$ with the shared
world-model transformer
\State Encode current and future observations independently with
frozen $E_\psi$; layer-normalize the resulting patch features
\State Project $\bm m$ and $s_t$ to $\tilde{\bm m}$ and $\tilde{\bm s}$
\State $\hat{\bm r}_{t+\Delta}\gets
G_\xi(\bm r_t;\tilde{\bm m},\tilde{\bm s})$
\State $\mathcal L_{\mathrm{RD}}\gets
\mathrm{SmoothL1}(\hat{\bm r}_{t+\Delta},\bm r_{t+\Delta})$
\State Backpropagate the weighted sum through trainable components
\end{algorithmic}
\end{algorithm}

\subsection{Action Inference and Optional Visualization}
\label{sec:appendix:infer}
Action inference requires the understanding path and action head only.
The deployed action branch uses 20 conditional Euler updates and
does not evaluate a null-language branch for classifier-free guidance.
The CDE and auxiliary future-generation branch are not needed for
action prediction.

\begin{algorithm}[H]
\caption{PILOT conditional action inference}\label{alg:infer}
\begin{algorithmic}[1]
\Require Current observation $o_t$, instruction $\ell$, state $s_t$,
$N=20$
\State $\bm c\gets\mathrm{WM}^{\mathrm{enc}}_\theta(o_t,\ell)$
\State $\bm a^{(0)}\sim\mathcal N(\bm0,\bm I)$
\For{$n=0,\ldots,N-1$}
\State $\tau\gets n/N$
\State $\hat{\bm v}\gets
\mathrm{ActionModel}_\phi(\bm a^{(n)},\tau,\bm c,s_t)$
\State $\bm a^{(n+1)}\gets\bm a^{(n)}+\hat{\bm v}/N$
\EndFor
\State \Return $\bm a^{(N)}$
\end{algorithmic}
\end{algorithm}

With future-video decoding disabled, the action-only inference path
bypasses the generation pipeline. Future videos can optionally be
generated for visualization, at additional denoising and VAE decoding cost.
Historical evaluation commands override its visual denoising count
to 5, without changing the action head's 20 Euler updates.
Action-only latency must therefore be measured with visual generation
explicitly disabled.

\FloatBarrier
\subsection{Separate Understanding and Generation Pipelines}
\label{sec:appendix:pipeline}
Algorithms~\ref{alg:understanding} and~\ref{alg:generation} separate
the two uses of the shared Cosmos transformer. The former returns
context tokens for Algorithm~\ref{alg:infer}; the latter returns a
video for optional visualization, not an input required by the action
head. The pseudocode specializes to one $224\times224$ observation.
The text encoder, VAE, and transformer parameters are shared between the two pipelines.

\begin{algorithm}[H]
\caption{Understanding pipeline: single-pass context extraction}
\label{alg:understanding}
\begin{algorithmic}[1]
\Require Current observation $o_t$, instruction $\ell$
\State $\bm p\gets\mathrm{TextEncoder}(\ell)$ \Comment{Frozen; conditional only}
\State $o\gets\mathrm{Preprocess}_{224}(o_t)$
\State $\bm z_{\rm obs}\sim q_{\rm VAE}(\bm z\mid o)$ \Comment{Frozen VAE}
\State $\bm z_{\rm obs}\gets(\bm z_{\rm obs}-\bm\mu)/\bm\sigma$
\State Set the observation condition mask to one and padding mask to zero
\State Set the conditional-frame timestep to $0.1$
\State Run the shared Cosmos transformer once on $(\bm z_{\rm obs},\bm p)$
\State $\bm c\gets$ last-block hidden states, reshaped to $196\times2048$
\State \Return $\bm c$ \Comment{No scheduler, future noise, CFG, or VAE decode}
\end{algorithmic}
\end{algorithm}

\begin{algorithm}[H]
\caption{Generation pipeline: optional future-video visualization}
\label{alg:generation}
\begin{algorithmic}[1]
\Require Current observation $o_t$, instruction $\ell$, visual steps $N_g$,
guidance $g$, output frame count $F$
\State Encode $\ell$ as $\bm p$; encode the negative prompt as $\bm p_-$ if $g>1$
\State Preprocess and pad the conditioning clip to $F$ frames
\State Sample and normalize VAE latents $\bm z_{\rm cond}$ of that clip
\State Construct condition mask $\bm M$ and temporal indicator $\bm I$ for the observed frame
\State $\bm z^{(0)}\sim\mathcal N(\bm0,\bm I)$
\State Initialize UniPC timesteps $\{t_i\}$ and noise levels $\{\sigma_i\}$
\State $\bm v_{\rm cond}\gets
           \bm M\odot(\bm z^{(0)}-\bm z_{\rm cond})$
\For{$i=0,\ldots,N_g-1$}
\State $\bm x_i\gets\bm M\odot\bm z_{\rm cond}
                         +(1-\bm M)\odot\bm z^{(i)}$
\State $\bm t_i\gets0.1\bm I+(1-\bm I)\sigma_i$
\State $\bm v_+\gets\bm v_{\rm cond}+(1-\bm M)\odot
          \mathrm{WM}_\theta(\bm x_i,\bm t_i,\bm p;\bm M)$
\If{$g>1$}
\State $\bm v_-\gets\bm v_{\rm cond}+(1-\bm M)\odot
          \mathrm{WM}_\theta(\bm x_i,\bm t_i,\bm p_-;\bm M)$
\State $\bm v_i\gets\bm v_++g(\bm v_+-\bm v_-)$
\Else
\State $\bm v_i\gets\bm v_+$
\EndIf
\State $\bm z^{(i+1)}\gets
                 \mathrm{UniPCStep}(\bm z^{(i)},\bm v_i,t_i)$
\EndFor
\State Decode $\bm\sigma\odot\bm z^{(N_g)}+\bm\mu$ with the frozen VAE
\State \Return the decoded video or its selected future frame
\end{algorithmic}
\end{algorithm}

Algorithm~\ref{alg:generation} follows the locally inspected
Cosmos-Predict2.5 Diffusers pipeline. For image conditioning, the input
image is followed by zero-valued image tensors before preprocessing,
and only the first latent frame is marked as observed.
The velocity on observed positions is replaced by $\bm v_{\rm cond}$;
the transformer sees clean observed latents at every iteration.
The displayed guidance formula uses this implementation's
$\bm v_++g(\bm v_+-\bm v_-)$ convention, not
$\bm v_-+g(\bm v_+-\bm v_-)$. Guidance is confined to video generation;
the action branch has no CFG.
The released configuration uses $F=5$, $N_g=20$, and $g=7$,
with one conditioning latent frame, matching the supervised
current/future temporal construction.
This optional diagnostic visualization is disabled in the reported
59.75\% action-only benchmark; it is not a demonstrated long-video rollout.

\subsection{Inference Efficiency}
\label{sec:appendix:efficiency}
\paragraph{Operation-count comparison.}
The action-only path requires one understanding forward and 20
Action-Perceiver evaluations per chunk. Neither the VJEPA encoder nor
the CDE is needed for this path. Optional video generation adds an
iterative use of the shared world model and one VAE decoding call.
Table~\ref{tab:inference_efficiency} retains a separate pipeline-timing
comparison using Cosmos-Predict2.5 and lists the released checkpoint's
RoboCasa result for the action-only strategy.

\begin{table}[t]
\centering
\caption{Pipeline strategy comparison on RoboCasa using Cosmos-Predict2.5.
Latency is reported in milliseconds; frequency is calculated as
$1000/\mathrm{latency}$. The decoupled success rate is the released
step-340000 checkpoint's 717/1200 result. The latency values and
predict-then-act score are retained from a separate measurement,
not a paired evaluation of that checkpoint.}
\label{tab:inference_efficiency}
\small
\setlength{\tabcolsep}{6pt}
\begin{tabular}{lccc}
\toprule
Strategy & \makecell{RoboCasa\\success (\%)} &
\makecell{Latency\\(ms)} & \makecell{Frequency\\(Hz)}\\
\midrule
Predict-then-Act & 56.5 & 1329.95 & 0.75\\
Decoupled (Ours) & 59.75 & 111.60 & 8.96\\
\bottomrule
\end{tabular}
\end{table}

\paragraph{Cost decomposition.}
Let $C_{\rm enc}$ denote text/VAE preprocessing for understanding,
$C_{\rm U}$ its single transformer forward, and $C_{\rm A}$ one
action-head evaluation. Action inference has the approximate cost
\begin{equation}
 C_{\rm action}=C_{\rm enc}+C_{\rm U}+20C_{\rm A}+C_{\rm output}.
 \label{eq:app:action-cost}
\end{equation}
$C_{\rm output}$ includes action conversion and transfer.
Generation additionally incurs
\begin{equation}
 C_{\rm video}=C'_{\rm enc}
       +N_g\left(\kappa C_{\rm G}+C_{\rm scheduler}\right)
       +C_{\rm decode}+C_{\rm post},
 \quad \kappa=1+\mathbb I[g>1].
 \label{eq:app:video-cost}
\end{equation}
$C_{\rm G}$ is a transformer call on the video latent sequence.
It need not equal $C_{\rm U}$: the released understanding input
contains one latent frame, whereas optional five-frame generation
uses the observed and future latent frames.
When optional visualization is enabled, the wrapper first generates the video and then predicts the action chunk before returning the combined response.
The cost decomposition assumes one logical VAE decode per video
request, irrespective of internal tiling. It excludes exceptional
retries and assumes one selected sample.

\paragraph{Interpretation and measurement protocol.}
Training-time RD can supervise the action queries without invoking
the CDE or decoding future video at action-only inference.
In the Cosmos-Predict2.5 comparison in
Table~\ref{tab:inference_efficiency}, decoupled inference reduces latency
from 1329.95\,ms to 111.60\,ms, a 91.6\% reduction (11.9$\times$ speedup).
The released action-only checkpoint obtains 59.75\% RoboCasa success.
The retained 56.5\% predict-then-act score and the timing measurements
come from a separate experiment; they do not establish a matched-checkpoint
success improvement or a fresh latency measurement for this release.
In particular, backbone-call ratios are not runtime ratios: sequence
length, guidance, preprocessing, action decoding, and scheduling all
contribute.

The timing comparison in Table~\ref{tab:inference_efficiency} was
performed on a single NVIDIA H20 GPU.
Action-only measurement must explicitly disable future-image decoding;
the evaluation wrapper's enabled default otherwise measures a different
mode.

For latency $T_{\rm chunk}$ in milliseconds, the maximum serial
replanning rate is $1000/T_{\rm chunk}$ Hz.
Dividing $T_{\rm chunk}$ by the horizon gives an amortized per-action
cost, not a closed-loop control rate. The supplied RoboCasa client
predicts 16 actions and executes 12 before replanning; a physical
control frequency additionally depends on the execution schedule and
communication overhead. In our real-world setup, the policy server
runs on a cloud server cluster and communicates with a local client,
so network latency can fluctuate. We therefore report a latency-derived
inference frequency, using the same reciprocal-latency convention for
both strategies. This provides a consistent reporting convention,
not a measurement of end-to-end closed-loop control frequency.

\FloatBarrier
\subsection{Training Configurations Across Benchmarks}
\label{sec:appendix:training}
Table~\ref{tab:train_hyp} summarizes the training and inference
settings for LIBERO, RoboCasa, and real-world experiments. For RoboCasa,
the action horizon is $H=16$, and the evaluation client executes
12 actions before replanning.
The training future target is selected after the action window,
corresponding to $\Delta=16$ away from episode boundaries.
For RoboCasa, padded state and action dimensions are 64 and 32, respectively.
Each branch uses its own image preprocessing; a common ImageNet
normalization for all branches is not assumed.

\begin{table}[p]
\centering
\caption{Training and inference hyperparameters for the three benchmarks.
Image size refers to the world-model input. For RoboCasa, the absolute
minimum learning rate applies to all trainable groups, and video steps and
CFG describe optional visualization, not action guidance.
Global batch counts demonstrations per optimizer update, before
action-noise replication. A dash indicates that the released RoboCasa
configuration does not establish the historical GPU count or global batch.}
\label{tab:train_hyp}
\small
\setlength{\tabcolsep}{3pt}
\renewcommand{\arraystretch}{1.04}
\begin{tabular}{@{}lccc@{}}
\toprule
Parameter & LIBERO & RoboCasa & Real-World \\
\midrule
\multicolumn{4}{@{}l}{\textit{Data and interfaces}}\\
World-model image size & $224\times224$ & $224\times224$ & $224\times224$\\
Selected training views & Primary & Single ego view & 3-view\\
Video reading backend & \texttt{torchvision\_av} & Decord & Decord\\
Future offset $\Delta$ & 16 & 16 & 16\\
Action dimension & 7 & 32 & 32\\
State dimension & 8 & 64 & 64\\
Action horizon $H$ & 50 & 16 & 30\\
\midrule
\multicolumn{4}{@{}l}{\textit{Optimization and learning rates}}\\
Optimizer & AdamW & AdamW & AdamW\\
Adam $\beta_1,\beta_2$ & $(0.9,0.95)$ & $(0.9,0.999)$ & $(0.9,0.999)$\\
Weight decay & 0.1 & 0.1 & 0.1\\
Selected training step & 100,000 & 340,000 & 100,000\\
Configured warmup & 1,000 & 2,000 & 1,000\\
Minimum learning rate & $10^{-6}$ & $2\times10^{-7}$ & $10^{-7}$\\
Gradient clipping & 1.0 & 1.0 & 1.0\\
World-model peak LR & $10^{-5}$ & $10^{-5}$ & $10^{-5}$\\
Action-head peak LR & $10^{-4}$ & $5\times10^{-5}$ & $5\times10^{-5}$\\
CDE peak LR & $10^{-4}$ & $5\times10^{-5}$ & $5\times10^{-5}$\\
Motion/state projector LR & $10^{-4}$ & $5\times10^{-5}$ & $5\times10^{-5}$\\
\midrule
\multicolumn{4}{@{}l}{\textit{Batch, hardware, and checkpointing}}\\
GPUs & $32\times\mathrm{H20}$ & -- & $32\times\mathrm{H20}$\\
Per-device batch size & 16 & 14 & 16\\
Global batch size & 512 & -- & 512\\
Precision & BF16 mixed & BF16 mixed & BF16 mixed\\
Save interval (steps) & 10,000 & 10,000 & 10,000\\
\midrule
\multicolumn{4}{@{}l}{\textit{Flow and optional visualization}}\\
Video inference steps & 20 & 20 & 20\\
Action inference steps & 20 & 20 & 20\\
CFG scale & 1.5 & 7.0 & 5.0\\
Independent action repeats & 4 & 4 & 4\\
\bottomrule
\end{tabular}
\end{table}

\paragraph{Configuration scope.}
The layer equations above specialize to the supplied RoboCasa
implementation; benchmark-specific settings are listed in
Table~\ref{tab:train_hyp}. BF16 mixed precision does not imply that
every component operates exclusively in BF16.
The selected RoboCasa run uses seed 42, a planned budget of 400,000
updates, and checkpoint selection at update 340,000.
Its cosine schedule specifies an absolute minimum learning rate of
$2\times10^{-7}$, rather than a group-wise minimum-to-peak ratio.
The archived per-device batch size is 14; the released configuration
does not establish the historical number of training devices.

\paragraph{Continuation and checkpoint provenance.}
The reported RoboCasa result uses the original step-340000 weight
export, whose checkpoint identity is recorded with the public logs.
The export does not include the historical optimizer, scheduler,
random-number-generator state, or data cursor.
Later bounded fine-tuning/resume checks are separate runs and do not
inherit this checkpoint's 59.75\% benchmark result.
The warmup in Table~\ref{tab:train_hyp} is the archived configured value,
not evidence of an additional warmup at checkpoint selection.

\FloatBarrier

\begin{figure}[t]
\centering
\centerline{\includegraphics[width=\linewidth]{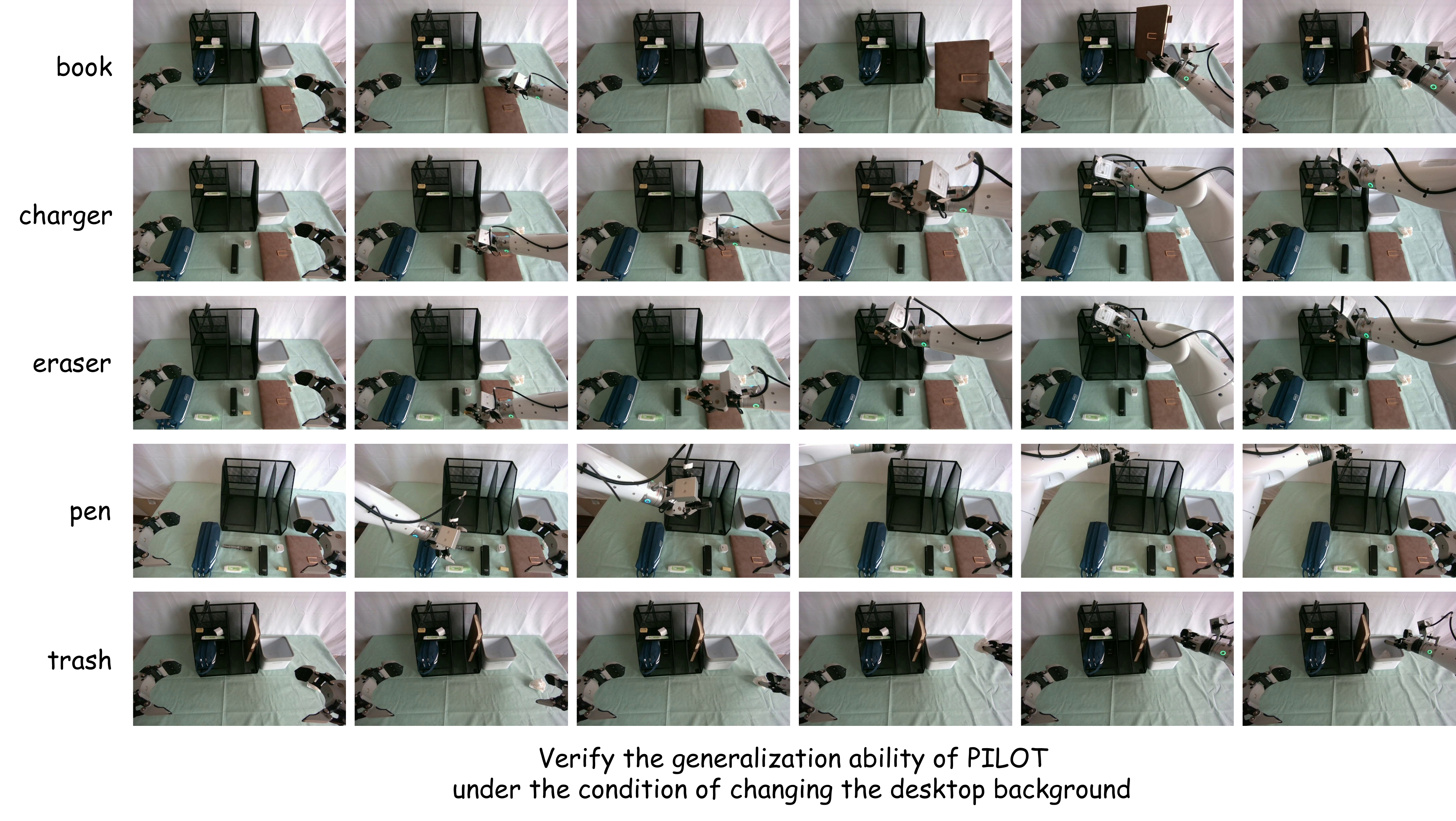}}
\caption{
Qualitative demonstrations under the \textit{Desk} setting, with a replaced desk surface and background. Multiple tasks are illustrated here; the quantitative Desk success rates in Table~\ref{tab:realworld} are evaluated on the Pen task only.
}\label{fig:append4}
\end{figure}

\begin{figure}[t]
\centering
\centerline{\includegraphics[width=\linewidth]{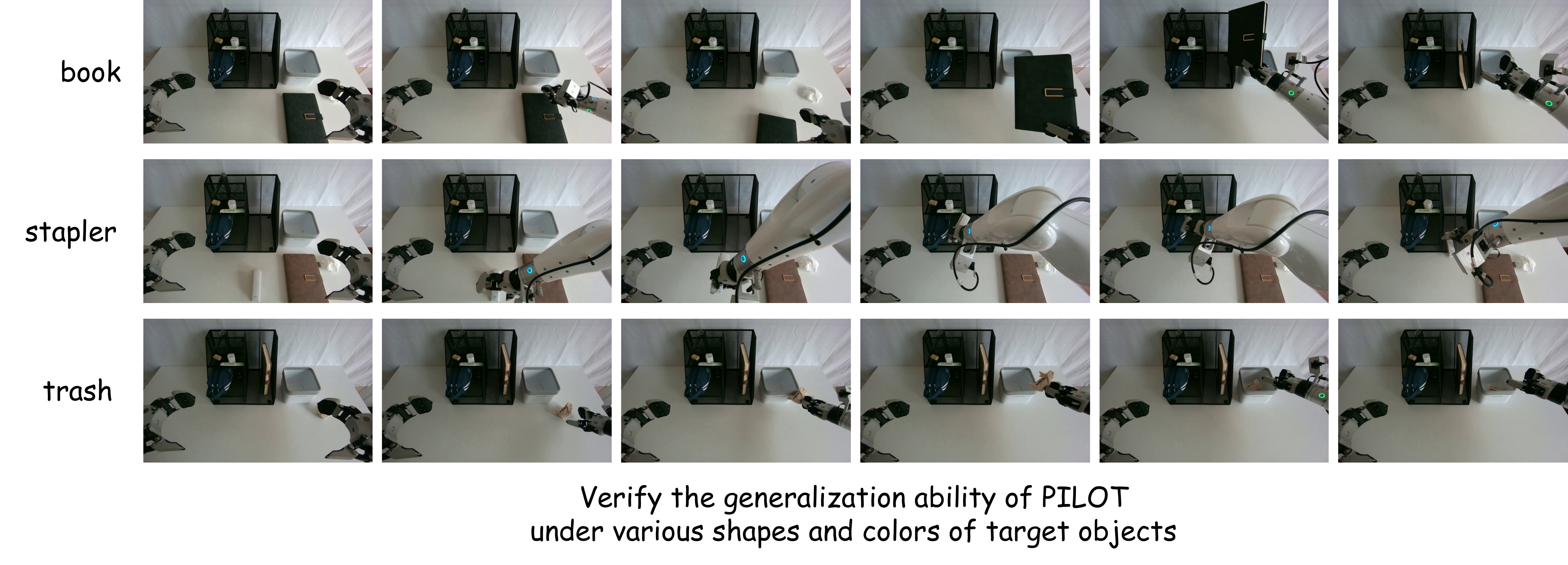}}
\caption{
Additional qualitative demonstrations with changed object shapes and colors. These examples involve Book, Stapler, and Trash, and are not the Pen-only quantitative Color evaluation reported in Table~\ref{tab:realworld}.
}\label{fig:append5}
\end{figure}

\subsection{Additional Experimental Details}
\label{sec:appendix:exp_details}

\paragraph{LIBERO benchmark.}
We evaluate on the Spatial, Object, Goal, and Long-horizon suites of
LIBERO~\citep{liu2023libero}. The LIBERO score is the average across
the four suite scores. LIBERO-Plus evaluates the standard-LIBERO-trained
policy under held-out perturbations, without training on the augmented
evaluation conditions~\citep{fei2025libero,zhang2026imagewam}.
The action space is 7-dimensional (EE pose + gripper), and the
proprioceptive state is 8-dimensional.
LIBERO training settings are summarized in Table~\ref{tab:train_hyp}.

\paragraph{RoboCasa-GR1 benchmark.}
We use the GR-1 Tabletop benchmark introduced with
GR00T~\citep{bjorck2025gr00tn1}, as implemented by the supplied
RoboCasa-GR1 environment. It is distinct from the single-arm
RoboCasa Kitchen benchmark in the same report.
Our task suite contains 24 manipulation tasks across 5 categories:
PnP Close (6), novel transfers from Cuttingboard (5), Placemat (4),
Tray (5), and Plate (4).
For the results in Tables~\ref{tab:robocasa} and~\ref{tab:robocasa_full},
we use the released step-340000 checkpoint and 50 episodes per task.
The overall success rate is 59.75\% (717/1200), averaging equally
over all 24 tasks.
The supplied mixture uses end-effector-pose data and the
\texttt{video.ego\_view} RGB observation, resized to
$224\!\times\!224$ for the world-model branch.
State and action vectors use per-dimension min-max normalization and
are padded to 64 and 32 dimensions, respectively.
The evaluation client removes unused output dimensions,
unnormalizes 27 active action values, and converts each arm's
6D end-effector pose to 7 joint commands through inverse kinematics.
Hand and waist commands are passed through their respective controls.
Thus, the padded action width is not the number of directly controlled
robot joints.

\paragraph{Baseline sources and method identities.}
GR00T-N1.6 follows Table~1 of StarVLA-$\alpha$~\citep{ye2026starvla}. LAPA$^*$ denotes the Prismatic-7B reimplementation reported in Table~I of UniVLA~\citep{bu2025univla}, not an original-paper LAPA evaluation; UniVLA uses the same table's full-pretraining row. OpenVLA-OFT, Mantis, VLA-JEPA, and the pretrained F1 variant use their respective reports~\citep{kim2025openvlaoft,yang2026mantisvla,sun2026vlajepa,lv2025f1}.
The original methods retain their own training and evaluation protocols.

LIBERO averages are computed across the four displayed suite scores.
For LIBERO-Plus, we recompute all aggregates from the displayed category scores, weighting Camera, Robot, Language, Light, Background, Noise, and Layout by $1599,1550,1537,1142,1076,1601,1525$, respectively (10{,}030 evaluation conditions in total)~\citep{fei2025libero}.
This applies a common weighting rule to all displayed rows. Since the
source category rates are rounded, the recomputed aggregates are
approximate summaries rather than reconstructions of raw success counts.

RoboCasa task-level results for GR00T-N1.5 and VP-VLA follow the VP-VLA report~\citep{wang2026vpvla}. GR00T-N1.6, UWM-XL, and LDA follow Table~VI of LDA~\citep{lyu2026lda}; UWM-XL is that paper's 1B-parameter UWM variant with Qwen3-VL encoding, not the original 140M-parameter UWM~\citep{zhu2025uwm}. DiT4DiT, PhysBrain-4B, and LangForce follow their respective reports~\citep{ma2026dit4dit,lin2025physbrain,lian2026langforce}.
The QwenGR00T column is the Qwen3-VL-based GR00T-style policy reported in the VP-VLA and PhysBrain comparisons, not the later StarVLA-$\alpha$ model. The PhysBrain column specifically denotes PhysBrain-4B.
In the full task table, baseline category and overall scores are
recomputed from the displayed task values and rounded to one decimal
place. PILOT aggregates are computed from the recorded success counts
and reported to two decimal places. The summary table additionally
retains the supplied TwinBrainVLA, StarVLA-$\alpha$, and ABot-M0 aggregates,
whose complete task-level columns are not included in the full table.
The overall average uses all 24 tasks, not an unweighted category mean.

The compared methods use distinct mechanisms.
$\pi_0$ and $\pi_{0.5}$ employ continuous-action flow matching~\citep{black2024pi0,intelligence2025pi05}.
GR00T N1 uses the Eagle-2 vision-language backbone~\citep{bjorck2025gr00tn1}.
OpenVLA-OFT denotes Optimized Fine-Tuning, with parallel continuous-action decoding and an L1 objective~\citep{kim2025openvlaoft}.
FAST compresses action chunks using a discrete cosine transform and tokenization~\citep{pertsch2025pi0fast}.
PokeVLA is a pocket-sized VLA with embodied-knowledge pretraining, goal-aware supervision, and geometry alignment~\citep{zheng2026pokevla}.
StarVLA-$\alpha$ studies a simplified VLA baseline with a pretrained VLM and a lightweight action head~\citep{ye2026starvla}.
VP-VLA uses visual prompting to communicate grounded plans to a VLA~\citep{wang2026vpvla}.
LangForce uses Bayesian decomposition and latent action queries to strengthen language grounding~\citep{lian2026langforce}.
FastWAM examines video co-training and the role of future imagination at inference~\citep{yuan2026fastwam}.

\paragraph{Evaluation protocol.}
The PILOT RoboCasa-GR1 evaluation in
Tables~\ref{tab:robocasa} and~\ref{tab:robocasa_full} uses 50 rollouts
per task, scene seeds 9000--9049 within each task, and a maximum of
720 executed physical steps.
Success is the native task predicate becoming true at any executed
physical step, including within an action chunk. Evaluation does not
terminate early on success, and the inverse-kinematics cache is cleared
at each episode reset. The released weights use legacy latent
coordinates, 20 action-sampling steps, and 12 executed actions per
16-action prediction; future-image generation is disabled.
RoboCasa category
scores average their constituent tasks; the overall score averages
all 24 tasks, equivalently weighting the five categories by
$6{:}5{:}4{:}5{:}4$.
The \href{https://github.com/pilot-wam-2026/pilot-code/blob/main/docs/LOGS.md}{public evaluation logs}
contain all 24 task result files and 1200 episode diagnostics.
They report 717 successes under this any-physical-step definition.
Scoring the same trajectories only at action-chunk endpoints gives
712 successes, and final-step-only scoring gives 637; these are
alternative scoring rules, not additional model runs.

\paragraph{Real-World benchmark.}
We deploy PILOT on an Agibot-G1 dual-arm humanoid robot to evaluate
its performance on real-world desk manipulation tasks.
The robot features 7-DOF arms on each side, a 1-DOF gripper per hand,
a 2-DOF head (pan/tilt), and a 2-DOF waist.
The policy interface uses a 32-dimensional action vector and a
64-dimensional proprioceptive state; these vector dimensions should
not be equated with the sum of the listed mechanical degrees of freedom.
Observations are obtained from three RGB cameras:
a head-mounted camera (\texttt{image\_head\_color}),
a left-hand wrist camera (\texttt{image\_hand\_left}),
and a right-hand wrist camera (\texttt{image\_hand\_right}),
providing complementary global and egocentric viewpoints.
The $224\!\times\!224$ resolution refers to preprocessing for the
world-model input, rather than native camera capture.
Figure~\ref{fig:append1} shows the Agibot-G1 robot, the standard evaluation scene, and illustrative changes to the target objects and desk surface.

\begin{figure}[H]
\centering
\centerline{\includegraphics[width=0.78\linewidth]{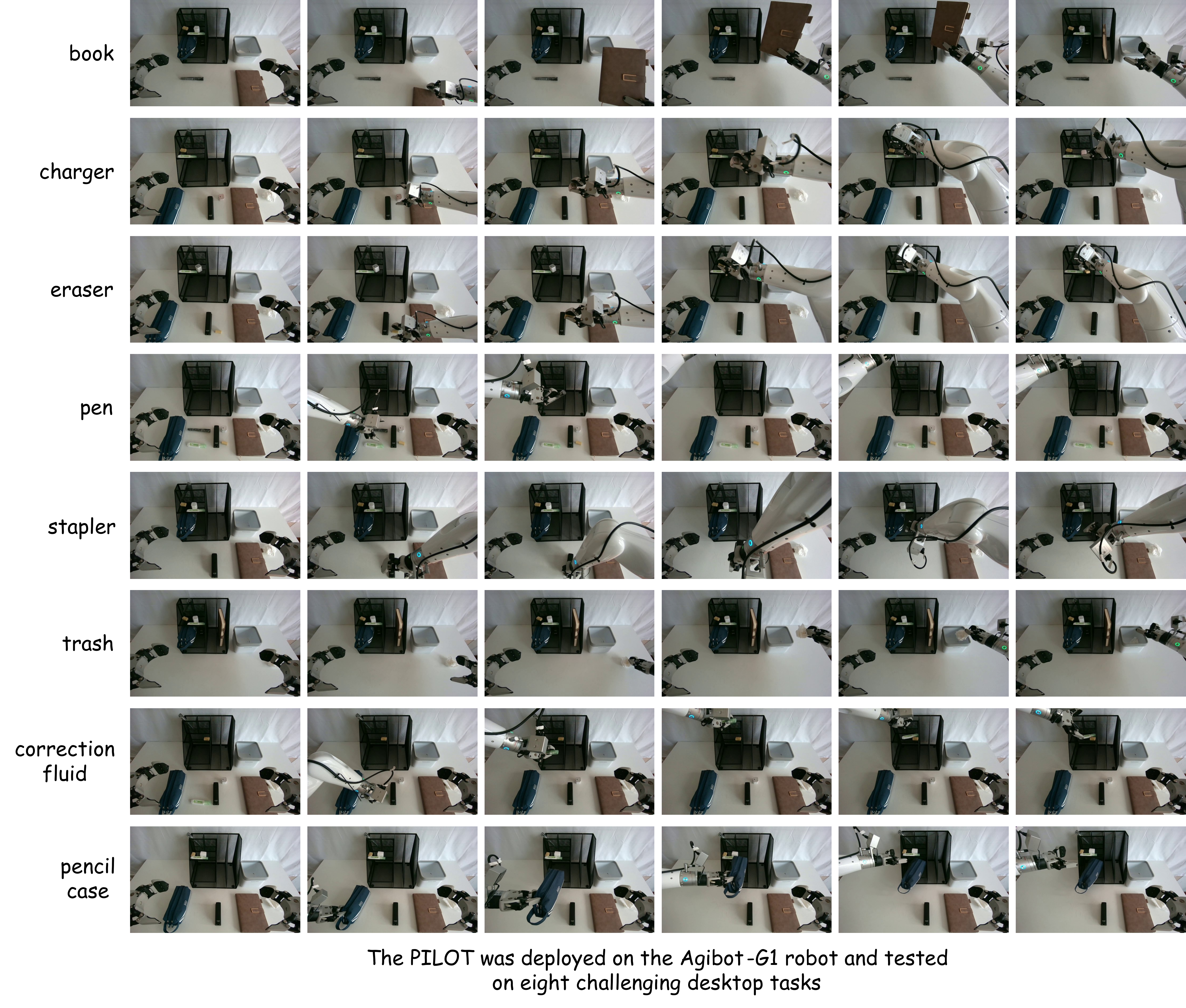}}
\caption{
PILOT deployed on the Agibot-G1 robot performing 8 standard desk manipulation tasks. The examples include inserting a pen into a grid slot, grasping and placing desktop objects, and placing a book upright into a storage rack.
}\label{fig:append2}
\end{figure}

The demonstration data is collected via teleoperation.
The \texttt{desk\_all\_3view} data mixture comprises 8
sub-datasets collected under consistent desk-top conditions,
each corresponding to a manipulation task with objects
commonly found on office desks.
In total, the mixture aggregates 32 source recording sessions,
comprising 31,720 demonstration trajectories across eight tasks,
with 3,965 trajectories per task and a combined total of 7.1M frames.
Visual observations are recorded at native capture resolutions of
$480\!\times\!848$ for the two wrist-mounted RealSense D405 cameras and
$800\!\times\!1280$ for the head-mounted camera.
All three cameras operate at 30\,fps.
The data is stored in the LeRobot format with min-max normalization
applied to all state and action keys.
We use the following 8 standard tasks for evaluation:
\textit{pen}, \textit{eraser}, \textit{correction fluid},
\textit{pencil case}, \textit{stapler},
\textit{book}, \textit{trash}, and \textit{charger}.
Figure~\ref{fig:append2} illustrates PILOT deployed on the Agibot-G1 robot performing the 8 standard tasks, including inserting a pen into a grid slot, grasping desktop objects, and picking up a book and placing it upright into a storage rack.

Table~\ref{tab:realworld_dataset} summarizes the dataset statistics
for each task.

\begin{table}[H]
\centering
\caption{Real-world desk manipulation dataset statistics.
\textit{Traj.} = number of demonstration trajectories;
\textit{Len.} = total frames divided by the reported trajectory count,
rounded to the nearest integer;
\textit{Frames} = total frames;
\textit{Freq.} = control frequency (Hz).
Trajectory counts are reported per task; the data mixture comprises
32 source recording sessions.}
\label{tab:realworld_dataset}
\begingroup
\normalsize
\tabcolsep=5pt
\renewcommand\arraystretch{1.0}
\noindent\begin{tabular*}{\linewidth}{@{\extracolsep{\fill}}lrrrr@{}}
\toprule
\textbf{Task} & \textbf{Traj.} & \textbf{Len.} & \textbf{Frames} & \textbf{Freq.} \\
\midrule
Pen            & 3,965  & 248  & 983,400  & 30 \\
Eraser         & 3,965  & 158  & 625,110  & 30 \\
Correction fluid & 3,965  & 220  & 872,097  & 30 \\
Pencil case    & 3,965  & 260  & 1,030,137  & 30 \\
Stapler        & 3,965  & 209  & 830,546  & 30 \\
Book           & 3,965  & 418  & 1,657,390  & 30 \\
Trash          & 3,965  & 135  & 537,195  & 30 \\
Charger        & 3,965  & 151  & 599,312  & 30 \\
\midrule
\textbf{Total} & \textbf{31,720} & --- & \textbf{7,135,187} & --- \\
\bottomrule
\end{tabular*}\par
\endgroup

\end{table}

\noindent\textit{Evaluation settings.}
We evaluate under three settings following the main text:
(1)~\textbf{Standard}: 50 rollouts per task with the default
initial state distribution;
(2)~\textbf{Generalization}: all three variants are evaluated on the \textit{Pen} task---\textit{Flash + Camera Offset},
which introduces sudden lighting changes and a shifted camera position;
\textit{Desk}, which replaces the desk surface and background;
and \textit{Color}, which replaces object colors with unseen variants;
Figures~\ref{fig:append3} and~\ref{fig:append4} provide additional
qualitative examples under lighting/viewpoint and desk changes.
Figure~\ref{fig:append5} illustrates object-appearance changes on other
tasks; it is not a record of the Pen-only Color success-rate experiment.

(3)~\blue{\textbf{Reduced-data (400 demonstrations)}: evaluated on the
\textit{Pen} task using the same fixed subset of 400 complete trajectories
for all methods. The protocol is detailed in
Appendix~\ref{sec:appendix:variability}; additional Book results are in
Appendix~\ref{sec:appendix:book}.}

\subsection{Additional Representational Analysis}
\raggedbottom
Figure~\ref{fig:pca_rgb} visualizes 1408-dimensional CDE representations
through a three-component PCA projection onto RGB channels. A
$16\times16$ patch grid describes spatial locations within one
representation, not an ordering of the $K$ Motion-CoT slots.
The overlays provide qualitative comparisons of spatial structure;
they are not calibrated attention maps or quantitative
prediction-error measurements.

\subsection{Additional Ablation Interpretation}\label{sec:appendix:ablation_interpretation}
Table~\ref{tab:ablation:component} reports component ablations on LIBERO.
Relative to the bare backbone, future-frame prediction and Motion-CoT alone add 2.3 and 3.2 percentage points.
Adding RD/CDE to Motion-CoT gives 1.6 points, followed by 0.8 from decoupled attention and 1.0 from future-frame prediction.
Decoupled attention blocks action-noise input to queries; the complete model retains all training objectives, while action prediction requires only the understanding--action pathway.
Because RD and CDE are enabled together in this table, their individual
effects are not isolated.

\clearpage
\subsection{Full RoboCasa-GR1 Results}
\label{sec:appendix:robocasa_results}

Table~\ref{tab:robocasa_full} gives all 24 tasks for PILOT and nine
baselines. PILOT obtains 717/1200 successes, or 59.75\%, with category
means of 47.67/70.00/54.50/73.20/53.50\% for
PnP Close/Cuttingboard/Placemat/Tray/Plate, respectively.
The corresponding success counts are
143/300, 175/250, 109/200, 183/250, and 107/200.
Within this ten-method table, PILOT leads Cuttingboard and Tray and
ties VP-VLA for first on Placemat. In the thirteen-method summary,
StarVLA-$\alpha$ leads Placemat and PILOT ties VP-VLA for second.
PILOT leads 11 individual tasks, including all five Tray tasks,
Cuttingboard-to-Pot (92\%), Cuttingboard-to-Pan (90\%), and
Plate-to-Plate (80\%). Placemat-to-Tieredshelf is second (30\%),
behind GR00T-N1.5 (36\%).

\begingroup
\renewcommand{\TableFont}{\small}
\begin{table}[H]
\centering
\caption{
\textbf{RoboCasa-GR1: full task results (\%)~\citep{bjorck2025gr00tn1}.}
PILOT uses the released step-340000 checkpoint and 50 rollouts per task (717/1200 successes). Averages use equal task weights; baseline sources are in Appendix~\ref{sec:appendix:exp_details}.
D4D: DiT4DiT; UWM: UWM-XL; QG: QwenGR00T; VP: VP-VLA; Phys.: PhysBrain-4B; Lang.: LangForce.
\colorbox{black!25}{Best} and \colorbox{black!10}{second-best} values are shaded per row, including ties.
}\label{tab:robocasa_full}
\TableFont
\setlength{\tabcolsep}{2pt}
\renewcommand{\arraystretch}{1.0}
\begingroup
\noindent\begin{tabular*}{\textwidth}{@{\extracolsep{\fill}}l|cccccccccc@{}}
\toprule
\textbf{Task}
& \makecell{GR00T\\N1.5}
& \makecell{GR00T\\N1.6}
& D4D
& UWM
& QG
& LDA
& VP
& Phys.
& Lang.
& PILOT \\
\midrule
\multicolumn{11}{l}{\textit{PnP Close}} \\
\midrule
PnP Bottle to Cab. & 54 & 51.5 & 48 & 41 & 46 & \best{76} & 54 & \second{74} & 72 & 64 \\
PnP Can to Drawer & 50 & 13 & 74 & 53 & \best{80} & 71 & 72 & 68 & \second{78} & 62 \\
PnP Cup to Drawer & 38 & 8.5 & \second{52} & 12 & \best{54} & 41 & 44 & 42 & 46 & 26 \\
PnP Milk to Micro. & \second{60} & 14 & 50 & 25 & 48 & 52 & \best{74} & 54 & 56 & 50 \\
PnP Potato to Micro. & 32 & \second{41.5} & 36 & 29 & 28 & 41 & 34 & 24 & 36 & \best{56} \\
PnP Wine to Cab. & 38 & 16.5 & 42 & 24 & 46 & \best{57} & 48 & \second{54} & 46 & 28 \\

\textbf{PnP Close (Avg)} & 45.3 & 24.2 & 50.3 & 30.7 & 50.3 & \best{56.3} & 54.3 & 52.7 & \second{55.7} & 47.67 \\
\midrule
\multicolumn{11}{l}{\textit{PnP Novel From Cuttingboard}} \\
\midrule
NvCB to Basket & 38 & 58 & 52 & 18 & 48 & 65 & \second{66} & 62 & \second{66} & \best{78} \\
NvCB to CbBox & 46 & 46.5 & 48 & 14 & 40 & \best{69} & \second{54} & 44 & 40 & 46 \\
NvCB to Pan & 58 & 68.5 & \second{76} & 20 & 68 & 75 & 74 & 56 & 68 & \best{90} \\
NvCB to Pot & 62 & \second{65} & 62 & 25 & 52 & 61 & 54 & 58 & 48 & \best{92} \\
NvCB to TBasket & 28 & 46.5 & 50 & 10 & \best{56} & \second{51} & \best{56} & 40 & 44 & 44 \\

\textbf{Cuttingboard * (Avg)} & 46.4 & 56.9 & 57.6 & 17.4 & 52.8 & \second{64.2} & 60.8 & 52.0 & 53.2 & \best{70.00} \\
\midrule
\multicolumn{11}{l}{\textit{PnP Novel From Placemat}} \\
\midrule
NvPM to Basket & 30 & \second{58.5} & 50 & 16 & 42 & 53 & 48 & 42 & 54 & \best{84} \\
NvPM to Bowl & 60 & 57.5 & 56 & 10 & 44 & 55 & \best{74} & 56 & \second{62} & 50 \\
NvPM to Plate & 56 & 63 & 32 & 12 & 48 & 59 & \second{70} & \best{80} & 52 & 54 \\
NvPM to TShelf & \best{36} & 28.5 & 18 & 2 & 18 & 24 & 26 & 14 & 24 & \second{30} \\

\textbf{Placemat * (Avg)} & 45.5 & \second{51.9} & 39.0 & 10.0 & 38.0 & 47.8 & \best{54.5} & 48.0 & 48.0 & \best{54.50} \\
\midrule
\multicolumn{11}{l}{\textit{PnP Novel From Tray}} \\
\midrule
NvTR to CbBox & 52 & 51.5 & 38 & 25 & 38 & \second{65} & 44 & 40 & 50 & \best{82} \\
NvTR to Plate & 48 & \second{71} & 56 & 18 & 56 & 63 & 66 & 66 & 58 & \best{80} \\
NvTR to Pot & 60 & \second{64.5} & 54 & 25 & 50 & 55 & 38 & 52 & 62 & \best{84} \\
NvTR to TBasket & 52 & 57 & 46 & 16 & 36 & 51 & \second{58} & 50 & 44 & \best{76} \\
NvTR to TShelf & 32 & 31.5 & \second{38} & 2 & 16 & 33 & 24 & 22 & 22 & \best{44} \\

\textbf{Tray * (Avg)} & 48.8 & \second{55.1} & 46.4 & 17.2 & 39.2 & 53.4 & 46.0 & 46.0 & 47.2 & \best{73.20} \\
\midrule
\multicolumn{11}{l}{\textit{PnP Novel From Plate}} \\
\midrule
NvPL to Bowl & \second{58} & 57 & 56 & 8 & \best{60} & 53 & 52 & 54 & 54 & 52 \\
NvPL to CbBox & 44 & 43.5 & \best{58} & 10 & \second{50} & 43 & 44 & \second{50} & 48 & 44 \\
NvPL to Pan & \second{60} & 51 & \best{68} & 20 & 54 & 55 & 56 & \best{68} & 54 & 38 \\
NvPL to Plate & 64 & \second{78.7} & 58 & 27 & 70 & 61 & 62 & 78 & 78 & \best{80} \\

\textbf{Plate * (Avg)} & 56.5 & 57.6 & \second{60.0} & 16.3 & 58.5 & 53.0 & 53.5 & \best{62.5} & 58.5 & 53.50 \\
\midrule
\textbf{Overall average} & 48.2 & 47.6 & 50.8 & 19.3 & 47.8 & \second{55.4} & 53.8 & 52.0 & 52.6 & \best{59.75} \\
\bottomrule
\end{tabular*}\par
\endgroup
\vspace{3pt}
{\TableFont\textbf{Abbreviations:} PnP = Pick and Place; Nv = Novel; CB = Cuttingboard; CbBox = Cardboardbox; PM = Placemat; TR = Tray; PL = Plate; TBasket = Tieredbasket; TShelf = Tieredshelf; Cab. = Cabinet; Micro. = Microwave.}

\end{table}

\endgroup

\end{document}